\documentclass[10pt,twocolumn,letterpaper]{article}

\usepackage{iccv}
\usepackage{times}
\usepackage{epsfig}
\usepackage{graphicx}
\usepackage{amsmath}
\usepackage{amssymb}

\usepackage{multirow}
\usepackage{booktabs}
\usepackage{makecell}
\usepackage{paralist}
\usepackage{subcaption}
\usepackage{pbox}
\usepackage[dvipsnames]{xcolor}
\usepackage{float}
\usepackage{comment}
\usepackage{authblk}
\usepackage{cleveref}
\usepackage[toc,page]{appendix}

\newcommand\blfootnote[1]{%
  \begingroup
  \renewcommand\thefootnote{}\footnote{#1}%
  \addtocounter{footnote}{-1}%
  \endgroup
}

\usepackage[pagebackref=true,breaklinks=true,letterpaper=true,colorlinks,bookmarks=false]{hyperref}

\iccvfinalcopy 

\def\iccvPaperID{1422} 

\ificcvfinal\fi
\begin{document}

\title{End-to-end Face Detection and Cast Grouping in Movies\\ Using Erd\H{o}s-R\'{e}nyi Clustering\vspace{-1cm}}
\author[1]{SouYoung Jin\vspace{-0.3cm}}
\author[1]{Hang Su}
\author[2]{Chris Stauffer}
\author[1]{Erik Learned-Miller}
\affil[1]{College of Information and Computer Sciences, University of Massachusetts, Amherst}
\affil[2]{Visionary Systems and Research (VSR)}

\twocolumn[{%
\renewcommand\twocolumn[1][]{#1}%
\maketitle
\thispagestyle{empty}
\begin{center}
	\vspace{-1.5cm}
    \setlength\tabcolsep{1pt}
    \setlength{\extrarowheight}{-3pt}
    \small{
    	\begin{tabular} {c c c c c}
          \includegraphics[width=0.2\textwidth]{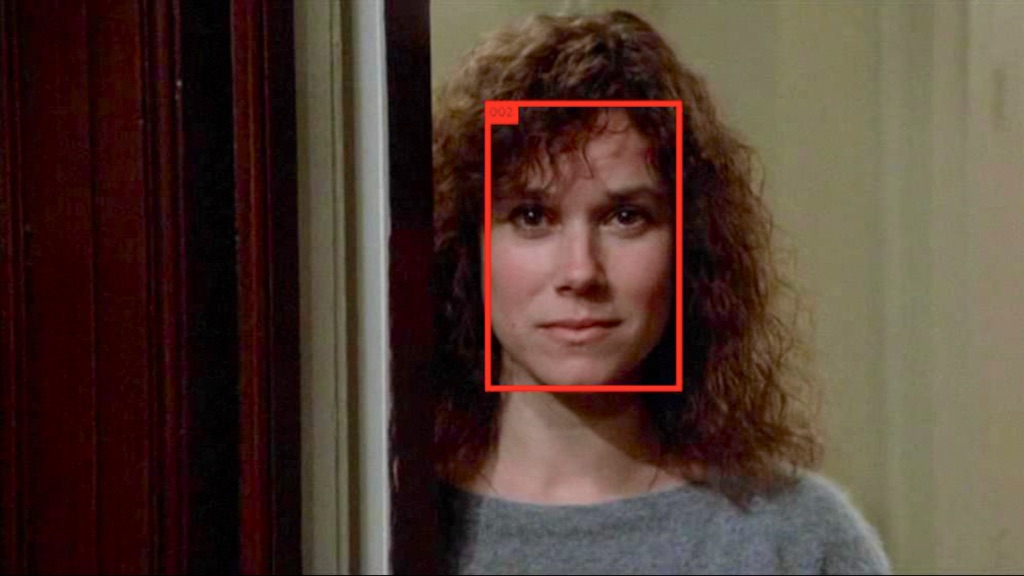} &
          \includegraphics[width=0.2\textwidth]{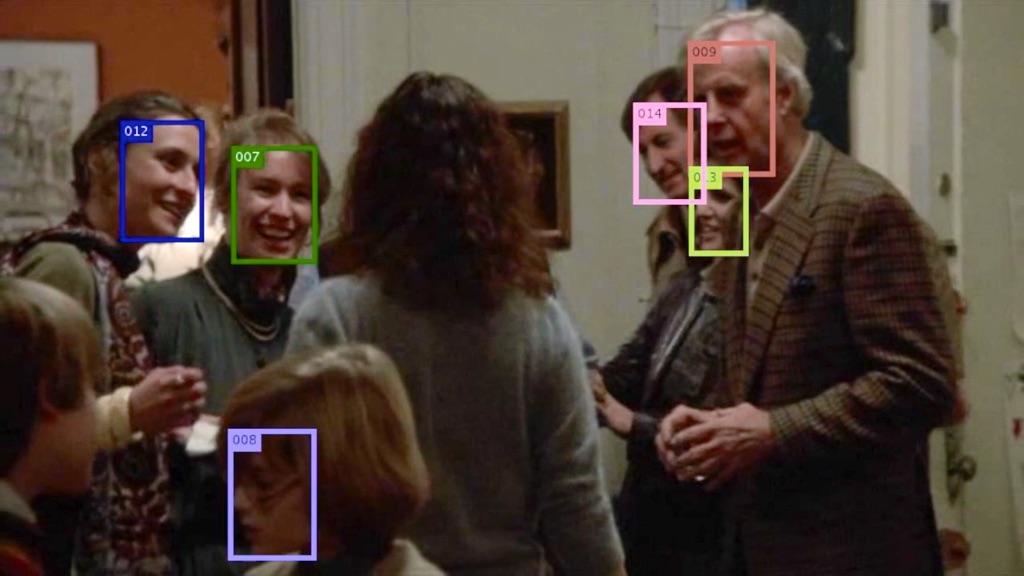} &
          \includegraphics[width=0.2\textwidth]{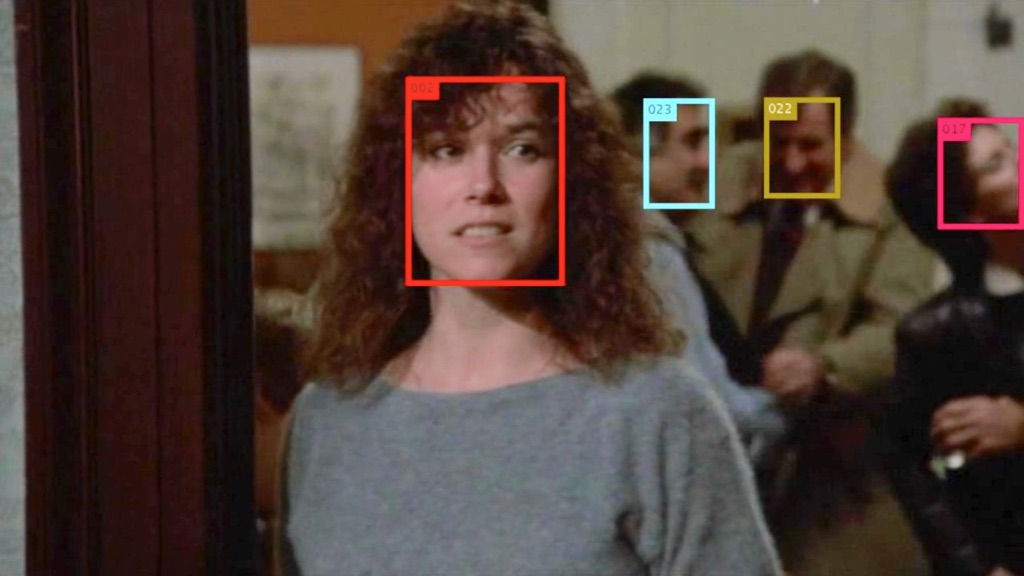} &
          \includegraphics[width=0.2\textwidth]{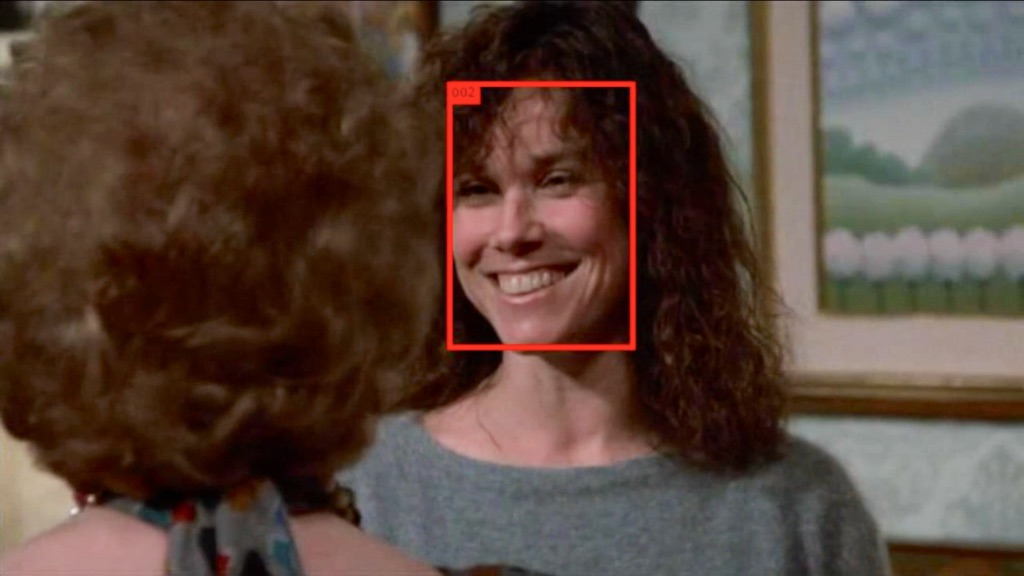} &
          \includegraphics[width=0.2\textwidth]{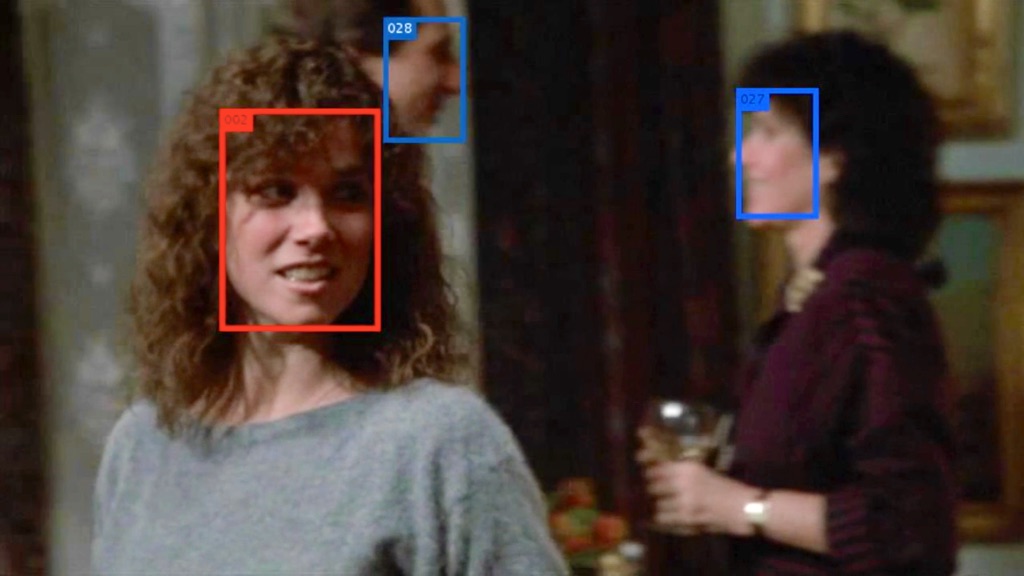}\\
          \includegraphics[width=0.2\textwidth]{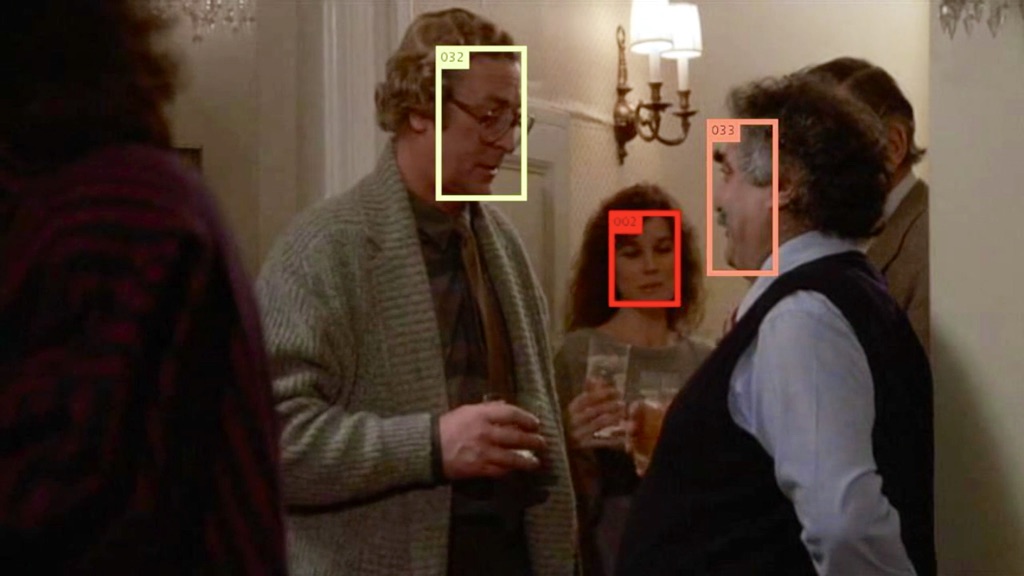} &
          \includegraphics[width=0.2\textwidth]{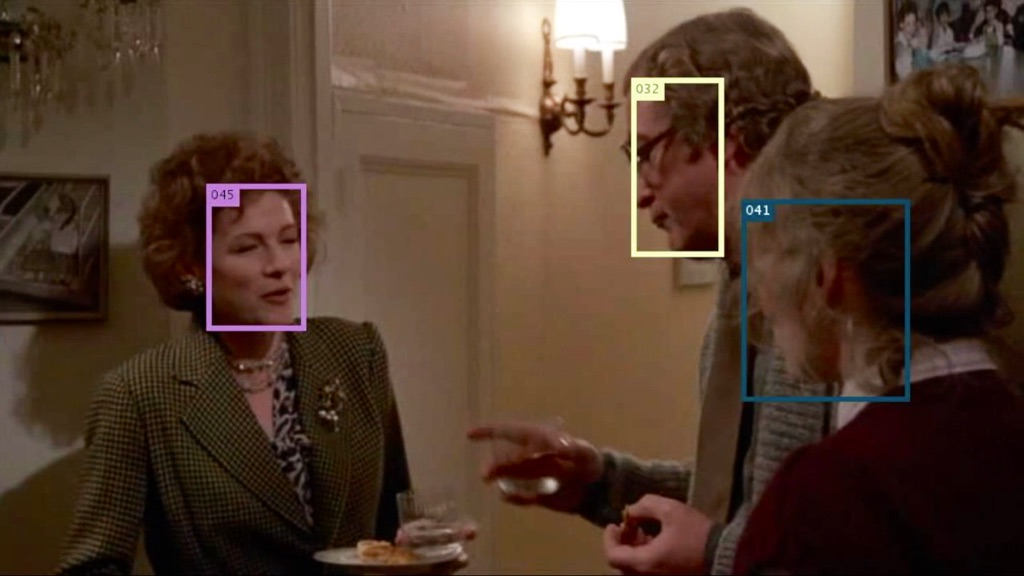} & 
          \includegraphics[width=0.2\textwidth]{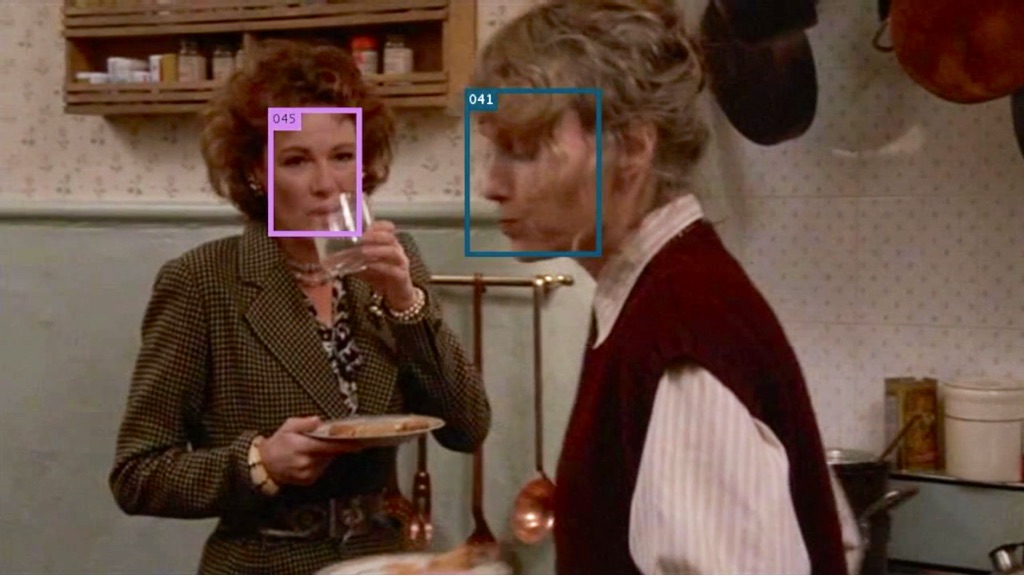} &
          \includegraphics[width=0.2\textwidth]{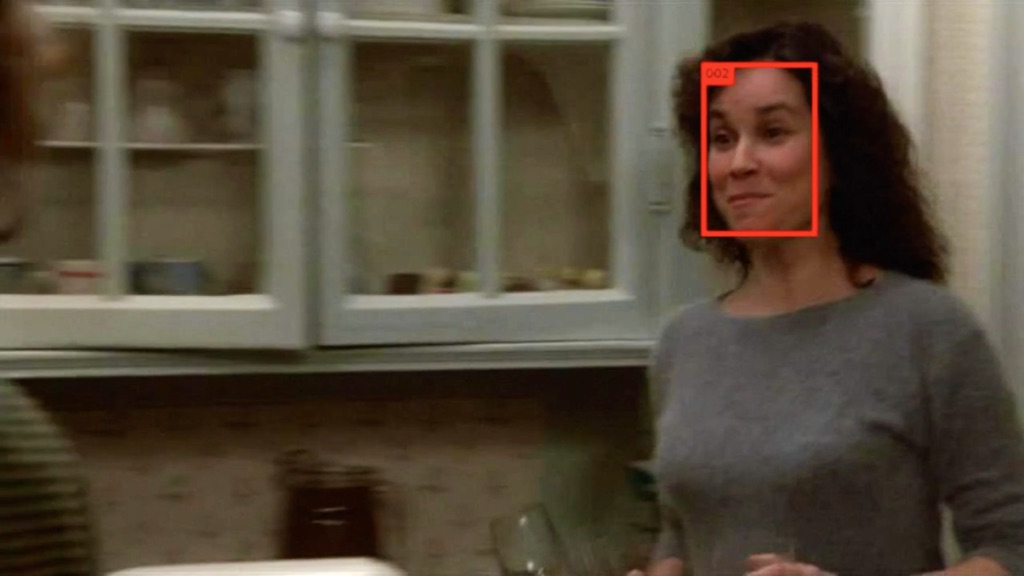} &
          \includegraphics[width=0.2\textwidth]{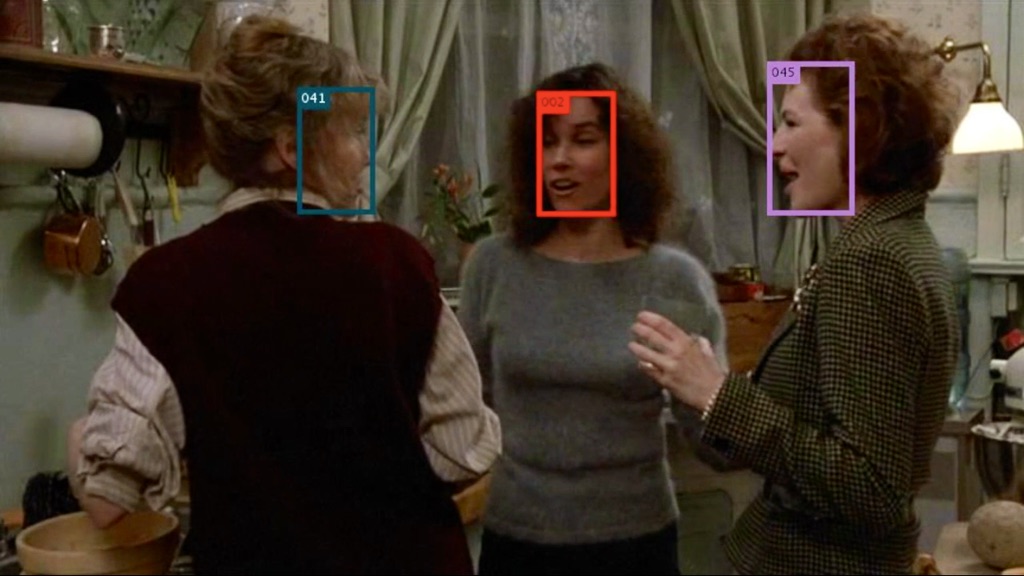}\\
      	\end{tabular}	
    }
    \vspace{-0.4cm}
    \captionof{figure}{{\bf Clustering results from \textit{Hannah and Her Sisters}.} 
    Each unique color shows a particular cluster. It can be seen that most individuals appear with a consistent color, indicating successful clustering.}
\end{center}
\label{fig:teaser}
}]
\begin{abstract}
\vspace{-0.4cm}
We present an end-to-end
system for detecting  and clustering faces by identity in full-length  movies. Unlike works that start with a predefined set of detected faces, we consider the end-to-end problem of detection and clustering together. We make three separate contributions. First, we combine a state-of-the-art face detector with a generic tracker to extract high quality face tracklets. We then introduce a novel clustering method, motivated by the classic graph theory results of Erd\H{o}s and R\'{e}nyi. It is based on the observations that  large clusters can be fully connected by joining just a small fraction of their point pairs, while just a single connection between two different people can lead to poor clustering results. This suggests clustering using a verification system with {\em very few false positives} but perhaps moderate recall. We introduce a novel verification method, {\bf rank-1 counts verification}, that has this property, and use it in a link-based clustering scheme.
Finally, we define a novel end-to-end detection and clustering evaluation metric allowing us to assess the accuracy of the entire end-to-end system.  We present state-of-the-art results on multiple video data sets and also on standard face databases.\blfootnote{This research is based in part upon work supported by the Office of the Director of National Intelligence (ODNI), Intelligence Advanced Research Projects Activity (IARPA) under contract number 2014-14071600010. The views and conclusions contained herein are those of the authors and should not be interpreted as necessarily representing the official policies or endorsements, either expressed or implied, of ODNI, IARPA, or the U.S. Government. The U.S. Government is authorized to reproduce and distribute reprints for Governmental purpose notwithstanding any copyright annotation thereon.}
\end{abstract}

\noindent
\vspace{-0.1cm}
{\bf Project page: \textcolor{blue}{http://souyoungjin.com/erclustering}}

\begin{figure*}[t]
\begin{center}
   \includegraphics[width=\linewidth]{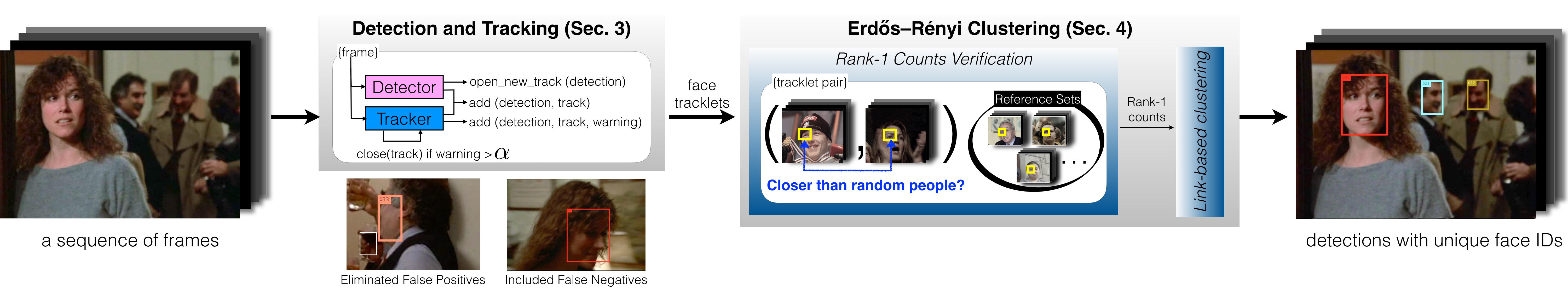}
\end{center}
   \vspace{-5mm}
   \caption{Overview of approach. Given a movie, our approach generates tracklets (Sec.~\ref{sec:overview}) and then does Erd\H{o}s-R\'{e}nyi Clustering and FAD verification between all tracklet pairs. (Sec.~\ref{sec:ErdosRenyi}) Our final output is detections with unique character Ids.}
\label{fig:overview_approach}
\end{figure*}

\section{Introduction}
The problem of identifying face images in video and clustering them
together by identity is a natural precursor to high impact
applications such as video understanding and analysis. This general
problem area was popularized in the paper ``Hello! My name
is...Buffy''~\cite{everingham2006cvpr}, which used text captions and face analysis to name people in each frame of a full-length video.  
In this work, we use only raw video (with no captions), and
group faces by identity rather than naming the characters. In addition, unlike face clustering methods that start with detected faces, we include detection as part of the problem. This means we must deal with false positives and false negatives, both algorithmically, and in our evaluation method. 
We make three contributions:
\begin{itemize}
\item A new approach to combining high-quality face detection~\cite{jiang2017} and generic tracking~\cite{lara2012cvpr} to improve both precision and recall of our video face detection.
\item A new method, {\em Erd\H{o}s-R\'{e}nyi clustering}, for large-scale clustering of images and video tracklets. We argue that effective large-scale face clustering requires face verification with fewer false positives, and we introduce {\em rank-1 counts verification}, showing that it indeed achieves better true positive rates in low false positive regimes. Rank-1 counts verification, used with simple link-based clustering, achieves high quality clustering results on three separate video data sets. 
\item A principled evaluation for the end-to-end problem of face detection and clustering in videos; until now there has been no clear way to evaluate the quality of such an end-to-end system, but only to evaluate its individual parts (detection and clustering). 
\end{itemize}

We structure the paper as follows. In Section~\ref{sec:relwork} we discuss related work. In Section~\ref{sec:overview}, we describe the first phase of our system, in which we use a face detector and generic tracker to extract face {\em tracklets}.  In Section~\ref{sec:ErdosRenyi}, we introduce Erd\H{o}s-R\'{e}nyi clustering and rank-1 counts verification. Sections~\ref{sec:experiments} and~\ref{sec:Discussion} present experiments and discussions.

\section{Related Work}
\label{sec:relwork}
In this section, we first discuss face tracking and then the problem of naming TV (or movie) characters. We can divide the character-naming work into two categories:  fully unsupervised and with some supervision. We then discuss prior work using reference images. Related work on clustering is covered in Section~\ref{sec:comparison}. 

Recent work on {\em robust face tracking}~\cite{tapaswi2014icip,roth2012icpr,Hannah} has gradually expanded the length of face tracklets, starting from face detection results. Ozerov~\etal~\cite{Hannah} merge results from different detectors by clustering based on spatio-temporal similarity. Clusters are then merged, interpolated, and smoothed for face tracklet creation. Similarly, Roth et al.~\cite{roth2012icpr} generate low-level tracklets by merging detection results, form high-level tracklets by linking low-level tracklets, and apply the Hungarian algorithm to form even longer tracklets. Tapaswi et al.~\cite{tapaswi2014icip} improve on this~\cite{roth2012icpr} by removing false positive tracklets.

With the development of multi-face tracking techniques, {\em the problem of naming TV characters}\footnote{Another related problem is {\em person re-identification}~\cite{zheng2016arxiv,lisanti2015pami,decann2015biometrics} in which the goal is to tell whether a person of interest seen in one camera has been observed by another camera. Re-identification typically uses the whole body on short time scales while {\em naming TV characters} focuses on faces, but over a longer period of time.
} 
has been also widely studied~\cite{tapaswi2012cvpr,haurilet2016cvpr,everingham2006cvpr,bauml2013cvpr,wu2013iccv,wu2013cvpr,tapaswi2015icvgip}. Given precomputed face tracklets, the goal is to assign a name or an ID to a group of face tracklets with the same identity. 
Wu et al.~\cite{wu2013iccv,wu2013cvpr} iteratively cluster face tracklets and link clusters into longer tracks in a bootstrapping manner. Tapaswi et al.~\cite{tapaswi2015icvgip} train classifiers to find thresholds for joining tracklets in two stages: within a scene and across scenes. 
Similarly, we aim to generate face clusters in a fully unsupervised manner.

Though solving this problem may yield a better result for face tracking, some forms of supervision specific to the video or characters in the test data can improve performance. Tapaswi et al.~\cite{tapaswi2012cvpr} perform face recognition, clothing clustering and speaker identification, where face models and speaker models are first trained on other videos containing the same main characters as in the test set. In~\cite{everingham2006cvpr,bauml2013cvpr}, subtitles and transcripts are used to obtain weak labels for face tracks. More recently, Haurilet et al.~\cite{haurilet2016cvpr} solve the problem without transcripts by resolving name references only in subtitles.
Our approach is more broadly applicable because it does not use subtitles, transcripts, or any other supervision related to the identities in the test data, unlike these other works~\cite{tapaswi2012cvpr,haurilet2016cvpr,everingham2006cvpr,bauml2013cvpr}.

As in the proposed verification system, some existing work~\cite{crosswhite2016arxiv,gyaourova2012tifs} uses reference images. For example, 
index code methods~\cite{gyaourova2012tifs} map each single
image to a code based upon a set of reference images, and then compare these codes. On the other hand, our method compares the relative distance of two images with the distance of one of the images to the reference set, which is different. In addition, we use the newly defined rank-1 counts, rather than traditional Euclidean or Mahalanobis distance measures to compare images~\cite{crosswhite2016arxiv,gyaourova2012tifs} for similarity measures.


\section{Detection and tracking}
\label{sec:overview}
Our goal is to take raw videos, with no captions or annotations, and to detect all faces and cluster them by identity. We start by describing our method for generating {\em face tracklets}, or continuous sequences of the same face across video frames. We wish to generate clean face tracklets that contain face detections from just a single identity. Ideally, exactly one tracklet should be generated for an identity from the moment his/her face appears in a shot until the moment it disappears or is completely occluded.

To achieve this, we first detect faces in each video frame using the {\em Faster R-CNN} object detector~\cite{FasterRCNN}, but retrained on the WIDER face data set~\cite{yang2015wider}, as described by Jiang et al.~\cite{jiang2017}. Even with this advanced detector, face detection sometimes fails under challenging illumination or pose. In videos, those faces can be detected before or after the challenging circumstances by using a tracker that tracks both forward and backward in time. We use the {\em distribution field tracker}~\cite{lara2012cvpr}, a general object tracker that is not trained specifically for faces. Unlike face detectors, the tracker's goal is to find in the next frame the object most similar to the target in the current frame. The extra faces found by the tracker compensate for missed detections (Fig.~\ref{fig:overview_approach}, bottom of block~2). Tracking helps not only to catch false negatives, but also to link faces of equivalent identity in different frames.

One simple approach to combining a detector and tracker is to run a tracker forward and backward in time from {\em every single face detection} for some fixed number of frames, producing a large number of ``mini-tracks". A Viterbi-style algorithm~\cite{forney1973viterbi, davey2008comparison} can then be used to combine these mini-tracks into longer sequences. This approach is computationally expensive since the tracker is run many times on overlapping subsequences, producing heavily redundant mini-tracks. To improve performance, we developed the following novel method for combining a detector and tracker. Happily, it also improves precision and recall, since it takes advantage of the tracker's ability to form long face tracks of a single identity. 

The method starts by running the face detector in each frame. When a face is first detected, a tracker is initialized with that face.  In subsequent frames, faces are again detected. In addition, we examine each current tracklet to see where it might be extended by the tracking algorithm in the current frame. We then check the agreement between detection and tracking results. We use the intersection over union (IoU) between detections and tracking results with  threshold 0.3, and apply the Hungarian algorithm\cite{kuhn1955hungarian} to establish correspondences among multiple matches. If a detection matches a tracking result, the detection is stored in the current face sequence such that the tracker can search in the next frame given the detection result. For the detections that have no matched tracking result, a new tracklet is initiated. If there are tracking results that have no associated detections, it means that either \textbf{a)} the tracker could not find an appropriate area on the current frame, or \textbf{b)} the tracking result is correct while the detector failed to find the face. The algorithm postpones its decision about the tracked region for the next $\alpha$ consecutive frames ($\alpha=10$). If the face sequence has any matches with detections within $\alpha$ frames, the algorithm will keep the tracking results. Otherwise, it will remove the tracking-only results. The second block of Fig.~\ref{fig:overview_approach} summarizes our proposed face tracklet generation algorithm and shows examples corrected by our joint detection-tracking strategy.
Next, we describe our approach to clustering based on low false positive verification.

\section{Erd\H{o}s-R\'{e}nyi Clustering and Rank-1 Counts Verification}
\label{sec:ErdosRenyi}
In this section, we describe our approach to clustering face images, or, in the case of videos, face tracklets. We adopt the basic paradigm of {\em linkage clustering}, in which each pair of points (either images or tracklets) is evaluated for linking, and then clusters are formed among all points connected by linked face pairs. We name our general approach to clustering {\em Erd\H{o}s-R\'enyi clustering} since it is inspired by  classic results in graph theory due to Erd\H{o}s and R\'enyi~\cite{ErdosRenyiGraphs}, as described next. 

\begin{figure}[ht]
\begin{center}
    \hspace*{-2.8mm}\includegraphics[width=1.12\linewidth]{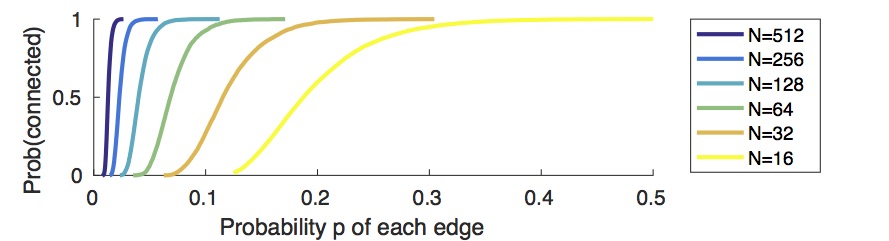}
\end{center}
\vspace{-2mm}
   \caption{Simulation of cluster connectedness as a function of cluster size, $N$, and the probability $p$ of connecting point pairs. The figure shows that for various $N$ (different colored lines), the probability that the cluster is fully connected (on the y-axis) goes up as more pairs are connected. For larger graphs, a small probability of connected pairs still leads to high probability that the graph will be fully connected.}
   \label{fig:simulation}
\end{figure}

Consider a graph $G$ with $n$ vertices and probability $p$ of each possible edge being present. This is the Erd\H{o}s-R\'{e}nyi random graph model~\cite{ErdosRenyiGraphs}. The expected number of edges is $\binom{n}{2}p.$
One of the central results of this work is that, for $\epsilon>0$ and $n$ sufficiently large, if 
\begin{equation}
\label{eq:ERG}
p>\frac{(1+\epsilon)\ln n}{n},
\end{equation}
then the graph will almost surely be connected (there exists a path from each vertex to every other vertex). Fig.~\ref{fig:simulation} shows this effect on different graph sizes, obtained through simulation.

Consider a clustering system in which links are made between tracklets 
by a {\em verifier} (a face verification system), whose job is to say whether a pair of tracklets is the ``same" person or two ``different" people.  While graphs obtained in clustering problems are not uniformly random graphs, the results of Erd\H{o}s and R\'{e}nyi suggest that this verifier can have a fairly low recall (percentage of 
same links that are connected) and still do a good job connecting large
clusters. In addition, false matches may connect large clusters of different identities, dramatically hurting clustering performance. This motivates us to build a verifier that focuses on low false positives rather than high recall. 
In the next section, we present our approach to building a verifier that is designed to have good recall at low false positive rates, and hence is appropriate for clustering problems with large clusters, like grouping cast members in movies.

\subsection{Rank-1 counts for fewer false positives}
\label{subsect:FAD}
Our method compares images by comparing their multidimensional feature vectors.
More specifically, we count the number of feature dimensions in which the two images are closer in value than the first image is to any of a set of reference images.
We call this number the {\em rank-1 count} similarity. Intuitively, two images whose feature values are ``very close" for many different dimensions are more likely to be the same person. Here, an image is considered ``very close" to a second image in one dimension if it is closer to the second image in that dimension than to any of the reference images. 

More formally, to compare two images $I_A$ and $I_B$, our first step is to obtain feature vectors $A$ and $B$ for these images. We extract 4096-D feature vectors from the \textit{fc7} layer of a standard pre-trained face recognition CNN~\cite{parkhi2015bmvc}. 
In addition to these two images, we  use a fixed reference set with $G$ images (we typically set $G=50$), and compute CNN feature vectors for each of these reference images.\footnote{The reference images may overlap in identity with the clustering set, but we choose reference images so that there is no more than one occurrence of each person in the reference set.} Let the CNN feature vectors for the reference images be $R^1,R^2,...,R^G$. We sample reference images from the {\em TV Human Interactions Dataset}~\cite{TV}, since these are likely to have a similar distribution to the images we want to cluster.

For each feature dimension $i$ (of the 4096), we ask whether 
$$ |A_i-B_i|< \min_j |A_i-R^j_i|.
$$
That is, is the value in dimension $i$ closer between $A$ and $B$ than between $A$ and all the reference images?  If so, then we say that the $i$th feature dimension  is {\em rank-1} between $A$ and $B$. The cumulative {\em rank-1 counts} feature $\mathbf{R}$ is simply the number of rank-1 counts across all 4096 features:
$$
\mathbf{R}=\sum_{i=1}^{4096} I\left[ |A_i-B_i|< \min_j |A_i-R^j_i|\right],
$$
where $I[\cdot]$ is an indicator function which is 1 if the expression is true and 0 otherwise.

Taking inspiration from Barlow's notion that the brain takes special note of ``suspicious coincidences" \cite{barlow1987cerebral}, each rank-1 feature dimension can be considered a suspicious coincidence. It provides some weak evidence that $A$ and $B$ may be two images of the same person. On the other hand, in comparing all 4096 feature dimensions, we expect to obtain quite a large number of rank-1 feature dimensions even if $A$ and $B$ are {\em not} the same person.  

When two images and the reference set are selected randomly from a large distribution of faces (in this case they are usually different people), the probability that $A$ is closer to $B$ in a particular feature dimension than to any of the reference images is just 
$$
\frac{1}{G+1}.
$$
Repeating this process 4096 times means that the expected number of rank-1 counts is simply
$$
E[\mathbf{R}]=\frac{4096}{G+1},
$$
since expectations are linear (even in the presence of statistical dependencies among the feature dimensions). Note that this calculation is a fairly tight {\em upper bound} on the expected number of rank-1 features {\em conditioned on the images being of different identities}, since most pairs of images in large clustering problems are different, and conditioning on "different" will tend reduce the expected rank-1 count. 
Now if two images $I_A$ and $I_B$ have a large rank-1 count, it is likely they represent the same person. The key question is how to set the threshold on these counts to obtain the best verification performance. 

\begin{figure}[!tb]
\begin{center}
    \includegraphics[width=0.9\linewidth]{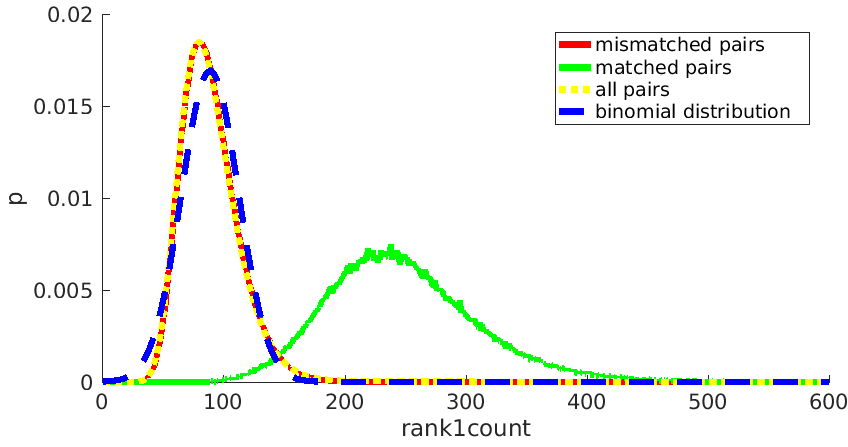}
    \vspace{-0.3cm}
\end{center}
\caption{LFW distribution of rank-1 counts. Each distribution is normalized to sum to 1.}
\label{fig:lfw_distribution}
\end{figure}

Recall that our goal, as guided by the Erd\H{o}s-R\'{e}nyi random graph model, is to find a threshold on the rank-1 counts $\mathbf{R}$ so that we obtain very few false positives (declaring two different faces to be ``same") while still achieving good recall (a large number of same faces declared to be ``same").  Fig.~\ref{fig:lfw_distribution} shows distributions of rank-1 counts for various subsets of image pairs from Labeled Faces in the Wild (LFW)~\cite{LFW}. The \textcolor{red}{red curve} shows the distribution of rank-1 counts for {\em mismatched} pairs from all possible mismatched pairs in the entire data set (not just the test sets). Notice that the mean is exactly where we would expect with a gallery size of 50, at $\frac{4096}{51}\approx 80$. The \textcolor{green}{green curve} shows the distribution of rank-1 counts for the matched pairs, which is clearly much higher. 
The challenge for clustering, of course, is that we don't have access to these distributions since we don't know which pairs are matched and which are not. The \textcolor{YellowOrange}{yellow curve} shows the rank-1 counts for {\em all} pairs of images in LFW, which is nearly identical to the distribution of mismatched rank-1 counts, {\em since the vast majority of possibe pairs in all of LFW are mismatched}. This is the distribution to which the clustering algorithm has access.

If the 4,096 CNN features were statistically independent (but not identically distributed), then the distribution of rank-1 counts would be a binomial distribution (\textcolor{blue}{blue curve}). In this case, it would be easy to set a threshold on the rank-1 counts to guarantee a small number of false positives, by simply setting the threshold to be near the right end of the mismatched (\textcolor{red}{red}) distribution. However, the dependencies among the CNN features prevent the mismatched rank-1 counts distribution from being binomial, and so this approach is not possible.

\subsection{Automatic determination of rank-1 count threshold}
Ideally, if we could obtain the rank-1 count distribution of mismatched pairs of a test set, we could set the threshold such that the number of false positives becomes very low. However, it is not clear how to get the actual distribution of rank-1 counts for mismatched pairs at test time.

Instead, we can estimate the shape of the mismatched pair rank-1 count distribution using one distribution (LFW), and use it to estimate the distribution of mismatched rank-1 counts for the test distribution. We do this by fitting the {\em left half} of the LFW distribution to the {\em left half} of the clustering distribution using scale and location parameters. The reason we use the left half to fit the distribution is that this part of the rank-1 counts distribution is almost exclusively influence by {\em mismatched pairs}. The {\em right side} of this matched distribution then gives us an approximate way to threshold the test distribution to obtain a certain false positive rate. It is this method that we use to report the results  in the leftmost column of Table~\ref{tab:clustering_comparison}.

\begin{table}[!tb]
\centering
\caption{Verification performance comparisons on all possible LFW pairs. The proposed rank-1 counts gets much higher recall at fixed FPRs.}
\label{tab:verification_comparison}
\scalebox{0.9}{
	\begin{tabular}{c | p{1cm} p{1cm} p{1cm} p{1cm} } 
        \Xhline{1.5pt}
        FPR
        & \textbf{ \rotatebox[origin=l]{90}{\pbox{5cm}{Rank1count}} }
        & \rotatebox[origin=l]{90}{\pbox{5cm}{L2}}
        & \rotatebox[origin=l]{90}{\pbox{5cm}{Template\\Adaptation~\cite{crosswhite2016arxiv}}} 
        & \rotatebox[origin=l]{90}{\pbox{5cm}{Rank-Order\\ Distance~\cite{zhu2011cvpr}}}
        \\
        \Xhline{1.5pt}
        1E-9 & \textbf{0.0252} & 0.0068 & 0.0016 & 0.0086 \\
        1E-8 & \textbf{0.0342} & 0.0094 & 0.0017 & 0.0086 \\
        1E-7 & \textbf{0.0614} & 0.0330 & 0.0034 & 0.0086 \\
        1E-6 & \textbf{0.1872} & 0.1279 & 0.0175 & 0.0086 \\
        1E-5 & \textbf{0.3800} & 0.3154 & 0.0767 & 0.0427 \\
        1E-4 & \textbf{0.6096} & 0.5600 & 0.2388 & 0.2589 \\
        1E-3 & 0.8222 & 0.7952 & 0.5215 & \textbf{0.8719} \\
        1E-2 & 0.9490 & 0.9396 & 0.8204 & \textbf{0.9656} \\
        1E-1 & \textbf{0.9939} & 0.9915 & 0.9776 & 0.9861 \\

        \Xhline{1.5pt}      
	\end{tabular}
}
\end{table}

A key property of our rank-1 counts verifier is that it has good recall across a wide range of the low false positive regime. Thus, our method is relatively robust to the setting of the rank-1 counts threshold. In order to show that our rank-1 counts feature has good performance for the types of verification problems used in clustering, we construct a verification problem using {\em all possible pairs} of the LFW database~\cite{LFW}. In this case, the number of mismatched pairs (quadratic in $N$) is much greater than the number of matched pairs. As shown in Table~\ref{tab:verification_comparison}, we observe that our verifier has higher recall than three competing methods (all of which use the same base CNN representation) at low false positive rates. 

{\bf Using rank-1 counts verification for tracklet clustering.} In our face clustering application, we consider every pair $(I,J)$ of tracklets, calculate a value akin to the rank-1 count $R$, and join the tracklets if the threshold is exceeded. In order to calculate an $R$ value for tracklets, we sample a random subset of 10 face images from each tracklet, compute a rank-1 count $R$ for each pair of images, and take the maximum of the resulting $R$ values. 

\subsection{Averaging over gallery sets}
While our basic algorithm uses a fixed (but randomly selected) reference gallery, the method is susceptible to the case in which one of the gallery images happens to be similar in appearance to a person with a large cluster, resulting in a large number of false negatives. To mitigate this effect, we implicitly average the rank-1 counts over an exponential number of random galleries, as follows.

The idea is to sample random galleries of size $g$ from a larger {\em super-gallery} with $G$ images; we used $g=50,G=1000$. We are interested rank-1 counts, in which image $A$'s feature is closer to $B$ than to any of the gallery of size $g$. Suppose we know that among the 1000 super-gallery images, there are $K$ (e.g., $K=3$) that are closer to $A$ than $B$ is. The probability that a random selection (with replacement) of $g$ images from the super-gallery would contain none of the $K$ closer images (and hence represent a rank-1 count) is
$$
r(A,B) = \Bigg( 1.0 - \frac{K}{G}\Bigg)^{g}.
$$
That is, $r(A,B)$ is the {\em probability} of having a rank-1 count with a random gallery, and using $r(A,B)$ as the count is equivalent to averaging over all possible random galleries. In our final algorithm, we sum these probabilities rather than the deterministic rank-1 counts.

\subsection{Efficient implementation}
\label{sec:time_complexity}
For simplicity, we discuss the computational complexity of our fixed gallery algorithm; the complexity of the average gallery algorithm is similar.
With $F$, $G$, and $N$ indicating the feature dimensionality, number of gallery images, and number of face tracklets to be clustered, the time complexity of the naive rank-1 count algorithm is $\mathcal{O}(F*G*N^2)$. 

However, for each feature dimension, we can sort $N$ test image feature values and $G$ gallery image feature values in time $\mathcal{O}((N+G) \log(N+G))$. Then, for each value in test image A, we find the closest gallery value, and increment the rank-1 count for the test images that are closer to A. Let $Y$ be the average number of stjpeg to find the closest gallery value. This is typically much smaller than $N$. The time complexity is then $\mathcal{O}(F*[(N+G) \log(N+G) + N*Y])$. 

\color{black}

\subsection{Clustering with do-not-link constraints}
\label{sec:donotlink}
It is common in clustering applications to incorporate constraints such as {\em do-not-link} or {\em must-link}, which specify that certain pairs should be in separate clusters or the same cluster, respectively~\cite{wagstaff2001icml, shental2004nips, lu2007neural, li2008icml, miyamoto2010fuzz}. They are also often seen in the face clustering literature~\cite{cinbis2011iccv, wu2013iccv, wu2013cvpr, ozerov2013icip, tapaswi2015icvgip, zhang2016eccv}. These constraints can be either rigid, implying they must be enforced~\cite{wagstaff2001icml, shental2004nips, miyamoto2010fuzz, ozerov2013icip}, or soft, meaning that violations cause an increase in the loss function, but those violations may be tolerated if other considerations are more important in reducing the loss~\cite{lu2007neural, li2008icml, wu2013iccv, wu2013cvpr, zhang2016eccv}. 

In this work, we assume that if two faces appear in the same frame, they must be from different people, and hence their face images obey a do-not-link constraint. Furthermore, we extend this hard constraint to the tracklets that contain faces. If two tracklets have any overlap in time, then the entire tracklets represent a do-not-link constraint.  

We enforce these constraints on our clustering procedure. Note that connecting all pairs below a certain dissimilarity threshold followed by transitive closure is equivalent to single-linkage agglomerative clustering with a joining threshold. In agglomerative clustering,  a pair of closest clusters is found and joined at each iteration until there is a single cluster left or a threshold met. A na\"\i ve implementation will simply search and update the dissimilarity matrix at each iteration, making the whole process $\mathcal{O}(n^3)$ in time. There are faster algorithms giving the optimal time complexity $\mathcal{O}(n^2)$ for single-linkage clustering~\cite{sibson1973slink, murtagh2012algorithms}. Many of these algorithms incur a dissimilarity update at each iteration, i.e.\ update $d(i, k) = \min(d(i, k), d(j, k))$ after combining cluster $i$ and $j$ (and using $i$ as the cluster id of the resulting cluster). If the pairs with do-not-link constraints are initialized with $+\infty$ dissimilarity, the aforementioned update rule can be modified to incorporate the constraints without affecting the time and space complexity: 
\[d(i, k) = \left\{
  \begin{array}{lr}
    \min(d(i, k), d(j, k)) & d(i, k) \neq +\infty \\
      & \text{AND}\; d(j, k) \neq +\infty \\
    +\infty & \text{otherwise}
  \end{array}
\right.
\]

\section{Experiments}
\label{sec:experiments}
We evaluate our proposed approach on three video data sets: {\em the Big Bang Theory} (BBT) Season~1 (s01), Episodes 1-6 (e01-e06)~\cite{bauml2013cvpr}, {\em Buffy the Vampire Slayer} (Buffy) Season 5 (s05), Episodes~1-6 (e01-e06)~\cite{bauml2013cvpr}, and {\em Hannah and Her Sisters} (Hannah)~\cite{Hannah}.
Each episode of the BBT and Buffy data set contains 5-8 and 11-17 characters respectively, while Hannah has annotations for 235 characters.\footnote{We removed garbage classes such as `unknown' or `false\_positive'.} Buffy and Hannah have many occlusions which make the face clustering problem more challenging. In addition to the video data sets, we also evaluate our clustering algorithm on LFW~\cite{LFW} which contains 5730 subjects.\footnote{All known ground truth errors are removed.}

\begin{table*}[!htb]
\centering
\caption{Clustering performance comparisons on various data sets. The leftmost shows our \textbf{rank1count} by setting a threshold automatically. For the rest of the columns, we show f-scores using optimal (oracle-supplied) thresholds. For BBT and Buffy, we show average scores over six episodes. The full table with individual episode results is given in Appendix~\ref{sec:performance_comparisons}. Best viewed in color (\textcolor{red}{\textbf{1st place}}, \textcolor{orange}{\textbf{2nd place}}, \textcolor{ForestGreen}{\textbf{3rd place}}).}
\label{tab:clustering_comparison}
\scalebox{0.8}{
	\begin{tabular}{ c | c | p{1.2cm}| p{1.2cm} p{1.2cm} p{1.2cm} p{1.2cm} | p{1.2cm} p{1.2cm} p{1.2cm} p{1.2cm} p{1.2cm} p{1.2cm} } 
        \Xhline{1.5pt}
        \multicolumn{2}{ c |}{ }
        		& \multicolumn{5}{ c |}{ Verification system + Link-based clustering algorithm} 
                & \multicolumn{6}{ c  }{ Other clustering algorithms }
                \\
        \cmidrule{3-13}
		\multicolumn{2}{ c |}{Test set}
        		& \textbf{ \rotatebox[origin=l]{90}{\pbox{5cm}{Rank-1 Count\\(automatic threshold)}} }
        		& \textbf{ \rotatebox[origin=l]{90}{\pbox{5cm}{Rank-1 Count}} }
                & \rotatebox[origin=l]{90}{\pbox{5cm}{L2}}
                & \rotatebox[origin=l]{90}{\pbox{5cm}{Template\\Adaptation~\cite{crosswhite2016arxiv}}} 
                & \rotatebox[origin=l]{90}{\pbox{5cm}{Rank-Order\\Distance~\cite{zhu2011cvpr}}}
                & \rotatebox[origin=l]{90}{\pbox{5cm}{Rank-Order Distance\\based Clustering~\cite{zhu2011cvpr}}}
                & \rotatebox[origin=l]{90}{\pbox{5cm}{Affinity\\Propagation~\cite{frey2007clustering}}} 
                & \rotatebox[origin=l]{90}{\pbox{5cm}{DBSCAN~\cite{ester1996density}}} 
                & \rotatebox[origin=l]{90}{\pbox{5cm}{Spectral\\Clustering~\cite{shi2000normalized}}} 
                & \rotatebox[origin=l]{90}{\pbox{5cm}{Birch~\cite{zhang1996birch}}} 
                & \rotatebox[origin=l]{90}{\pbox{5cm}{MiniBatch\\KMeans~\cite{sculley2010web}}} 
                \\
\Xhline{1.5pt}
\multirow{3}{*}{Video} 
& \pbox{2cm}{BBT s01 \cite{bauml2013cvpr}}
	& \textcolor{orange}{\textbf{.7728}} & \textcolor{red}{\textbf{.7828}} & .7365 & \textcolor{ForestGreen}{\textbf{.7612}} & .6692 & .6634 & .1916 & .2936 & .6319 & .2326 & .1945 \\
& \pbox{2cm}{Buffy s05 \cite{bauml2013cvpr}}
	& \textcolor{ForestGreen}{\textbf{.5661}} & \textcolor{red}{\textbf{.6299}} & .3931 & \textcolor{orange}{\textbf{.5845}} & .2990 & .5439 & .1601 & .1409 & .5351 & .1214 & .1143 \\
& Hannah~\cite{Hannah}
& \textcolor{orange}{\textbf{.6436}} & \textcolor{red}{\textbf{.6813}} & .2581 & .3620  & \textcolor{ForestGreen}{\textbf{.4123}} & .3955 & .1886 & .1230 & .3344 & .1240 & .1052 \\ 
\midrule                                
\multirow{1}{*}{Image} & \multicolumn{1}{ c |}{LFW~\cite{LFW}} 
& \textcolor{orange}{\textbf{.8532}} & \textcolor{red}{\textbf{.8943}} & \textcolor{ForestGreen}{\textbf{.8498}} & .3735 & .5989 & .5812 & .3197 & .0117 & .2538 & .4520 & .3133 \\ 
\Xhline{1.5pt}      
\end{tabular}
}
\end{table*}

{\bf An end-to-end evaluation metric}.
\begin{figure}[tb]
    \centering
	\scalebox{0.8}{
      \begin{subfigure}[b]{0.23\textwidth}
          \includegraphics[width=\textwidth,trim=0 -0mm 0 0,clip]{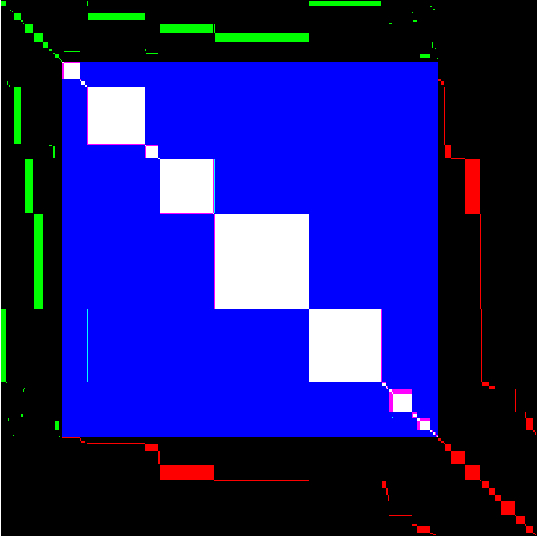}
          \caption{Rank-1 Count}
      \end{subfigure}
      \begin{subfigure}[b]{0.23\textwidth}
          \includegraphics[width=\textwidth]{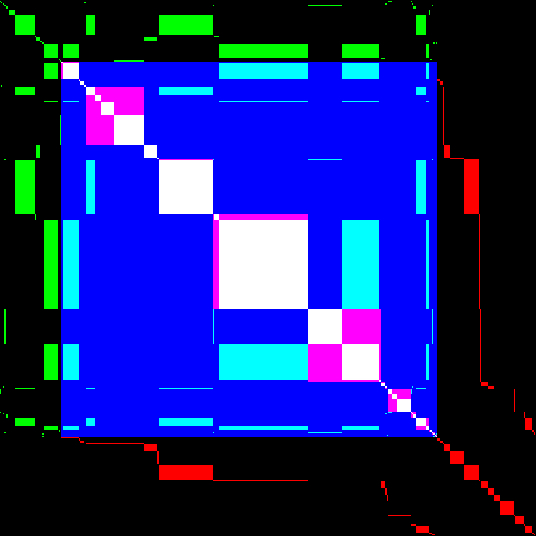}
          \caption{Rank-Order Distance~\cite{zhu2011cvpr}}
      \end{subfigure}    
      }
    \caption{Visualization of the combined detection and clustering metric for the first few minutes of the Hannah set.\vspace{-0.5cm}}
\label{fig:visualization_of_metric}
\end{figure}
There are many evaluation metrics used to independently evaluate detection, tracking, and clustering.  Previously, it has been difficult to evaluate the relative performance of two end-to-end systems because of the complex trade-offs between detection, tracking, and clustering performance.  Some researchers have attempted to overcome this problem by providing a reference set of detections with suggested metrics~\cite{MOT16}, but this approach precludes optimizing complete system performance.  To support evaluation of the full video-to-identity pipeline, in which false positives, false negatives, and clustering errors are handled in a common framework, we introduce {\em unified pairwise precision} (UPP) and {\em unified pairwise recall} (UPR) as follows.

Given a set of annotations, $\left \{a_1,a_2,...,a_A\right \}$ and detections, $\left \{d_1,d_2,...,d_D\right \}$, we consider the union of three sets of tuples: false positives resulting from unannotated face detections $\left \{d_i,\emptyset \right \}$; valid face detections $\left \{d_i,a_j \right \}$; and false negatives resulting from unmatched annotations $\left \{\emptyset,a_j \right \}$.  Fig.~\ref{fig:visualization_of_metric} visualizes every possible pair of tuples ordered by false positives, valid detections, and false negatives for the first few minutes of the Hannah data set.  Further, groups of tuples have been ordered by identity to show blocks of identity to aid our understanding of the visualization, although the order is inconsequential for the numerical analysis.

In Fig \ref{fig:visualization_of_metric}, the large blue region (and the regions it contains) represents all pairs of annotated detections, where we have valid detections corresponding to their best annotation.  In this region, white pairs are correctly clustered, \color{magenta}magenta\color{black}~pairs are the same individual but not clustered, \color{cyan}cyan\color{black}~pairs are clustered but not the same individual, and \color{blue}blue\color{black}~pairs are not clustered pairs from different individuals.  The upper left portion of the matrix represents false positives with no corresponding annotation.  The \color{green}green\color{black}~pairs in this region correspond to any false positive matching with any valid detection.  The lower right portion of the matrix corresponds to the false negatives.  The \color{red}red\color{black}~pairs in this region correspond to any missed clustered pairs resulting from these missed detections.  The ideal result would contain blue and white pairs, with no green, red, cyan, or magenta.

The unified pairwise precision (UPP) is the fraction of pairs, $\left \{d_i,a_j \right \}$ within all clusters with matching identities, \ie, the number of white pairs divided by the number of white, cyan, and green pairs.  UPP decreases if: two matched detections in a cluster do not correspond to the same individual; if a matched detection is clustered with a false positive; for each false positive regardless of its clustering; and for false positives clustered with valid detections.  Similarly, the unified pairwise recall (UPR) is the fraction of pairs within all identities that have been properly clustered, \ie, the number of white pairs divided by number of white, magenta, and red pairs.  UPR decreases if: two matched detections of the same identity are not clustered; a matched detection should be matched but there is no corresponding detection; for each false negative; and for false negative pairs that should be detected and clustered.  The only way to achieve perfect UPP and UPR is to detect every face with no false positives and cluster all faces correctly.  At a glance, our visualization in Fig.~\ref{fig:visualization_of_metric} shows that our detection produces few false negatives, many more false positives, and is less aggressive in clustering.  Using this unified metric, others can tune their own detection, tracking, and clustering algorithms to optimize the unified performance metrics. Note that for image matching without any detection failures, the UPP and UPR reduce to standard pairwise precision and pairwise recall.

The UPP and UPR can be summarized with a single F-measure (the weighted harmonic mean) providing a single, unified performance measure for the entire process. It can be $\alpha$-weighted to alter the relative value of precision and recall performance:
\begin{equation}
F_\alpha={{1}\over{{\alpha\over{UPP}} + {{1-\alpha}\over{UPR}}}}
\end{equation}
where $\alpha \in [0,1]$. $\alpha=0.5$ denotes a balanced F-measure.

\begin{figure*}[h]
	\centering
	\setlength\tabcolsep{1pt}
    \setlength{\extrarowheight}{-3pt}
    \small{
      \begin{tabular} {c c c c c}
      \includegraphics[width=0.2\textwidth]{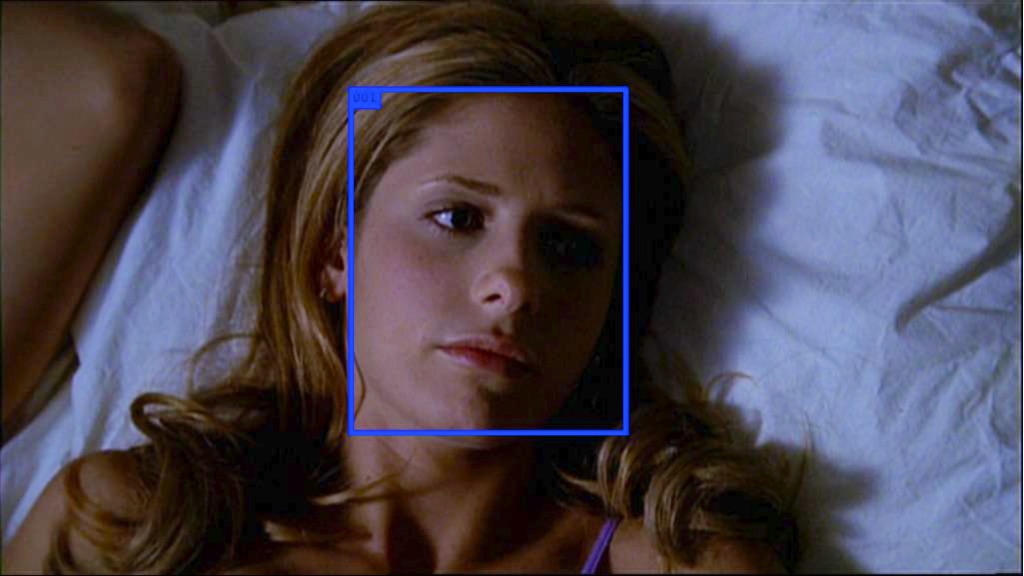} &
      \includegraphics[width=0.2\textwidth]{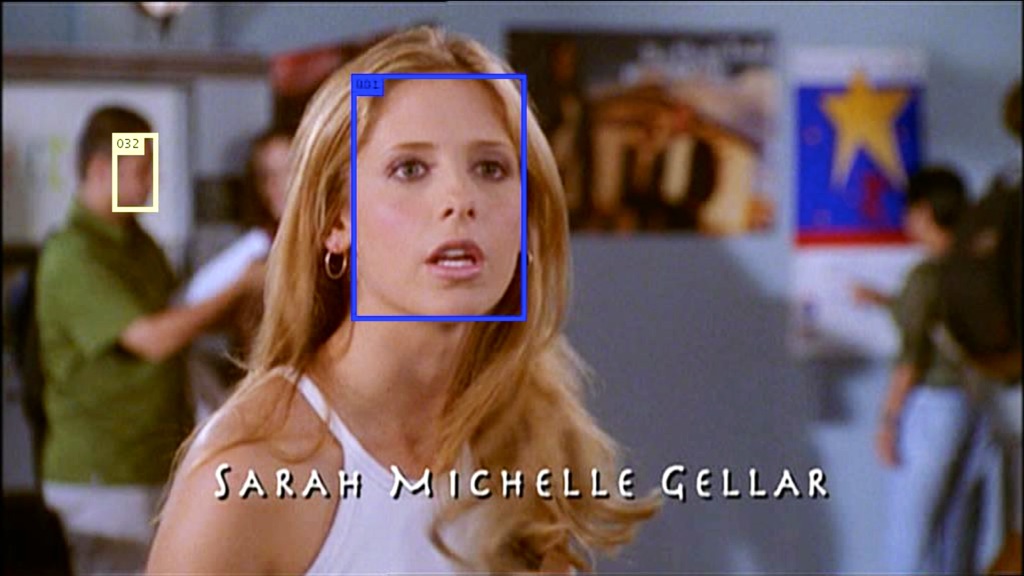} &
      \includegraphics[width=0.2\textwidth]{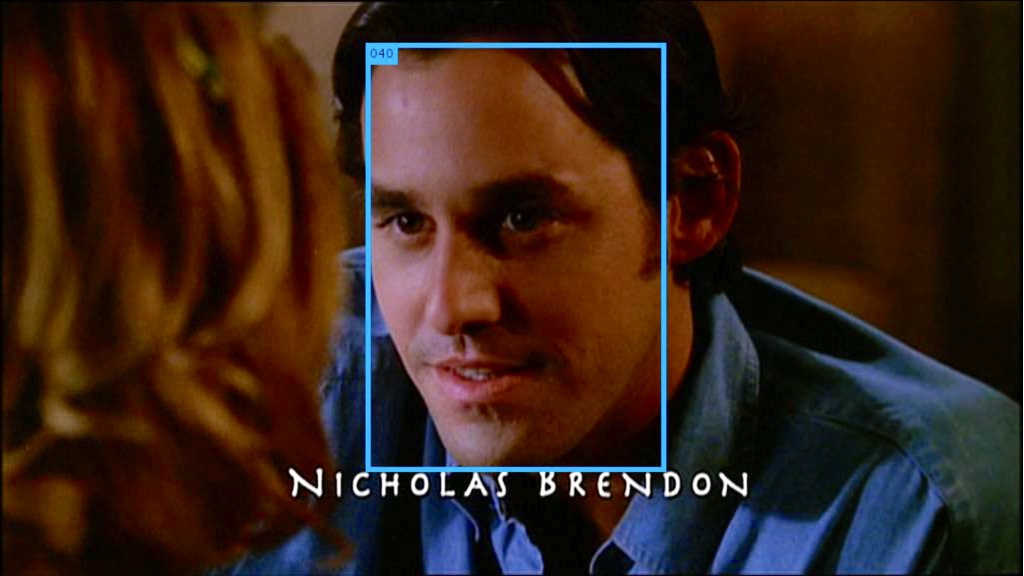} &
      \includegraphics[width=0.2\textwidth]{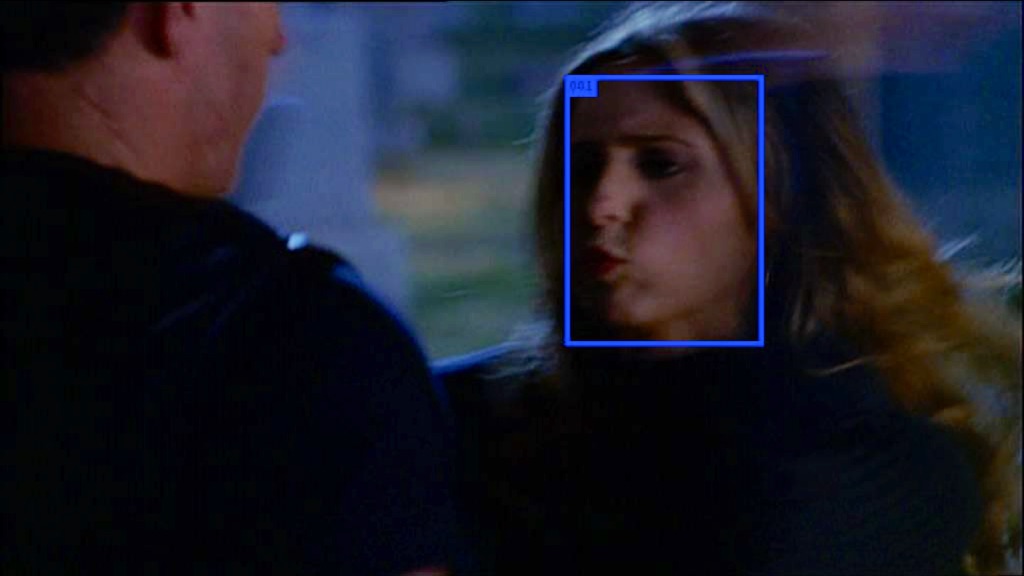} &
      \includegraphics[width=0.2\textwidth]{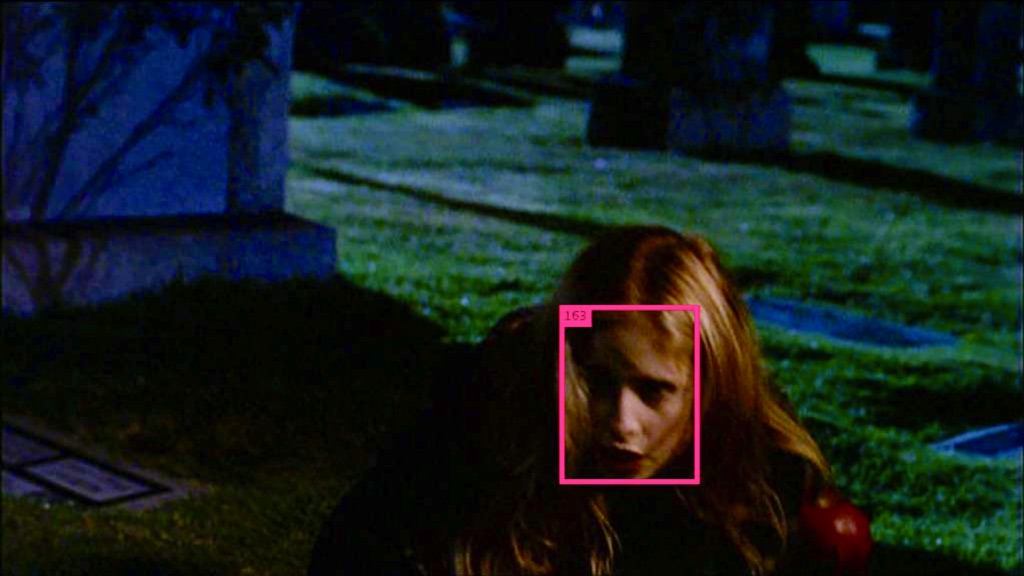} \\
      (a) & (b) & (c) & (d) & (e)\\
      \includegraphics[width=0.2\textwidth]{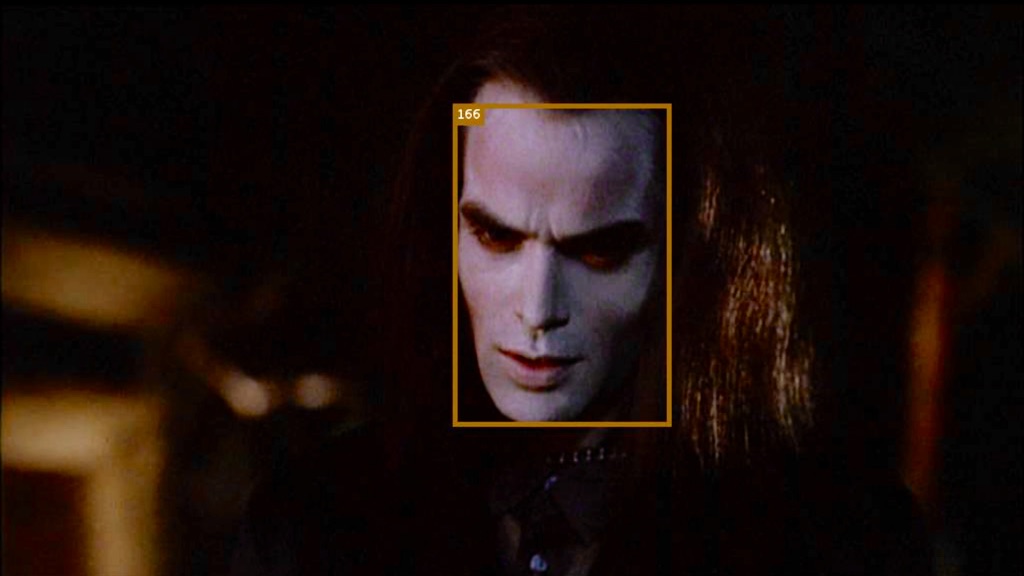} &
      \includegraphics[width=0.2\textwidth]{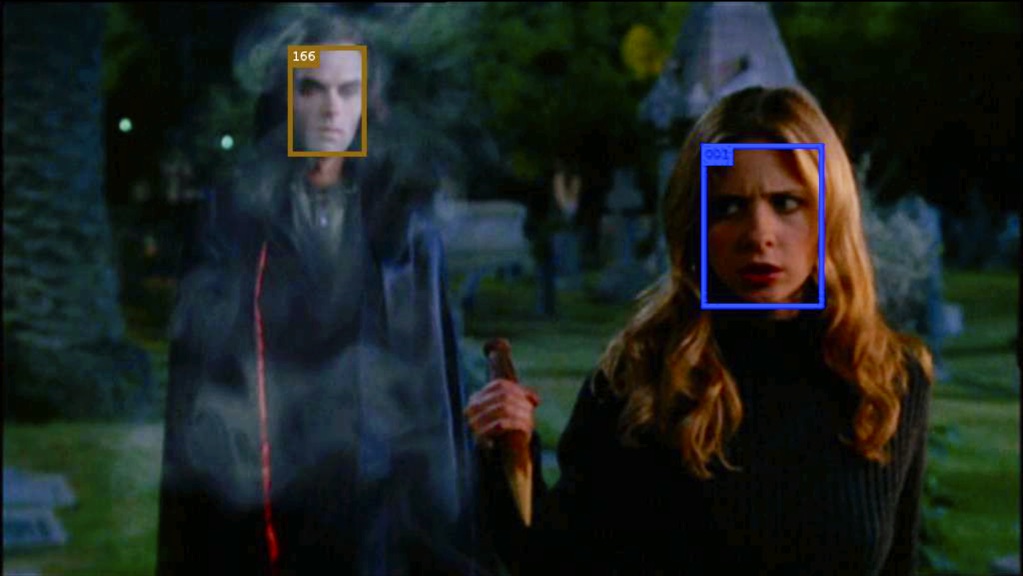} &
      \includegraphics[width=0.2\textwidth]{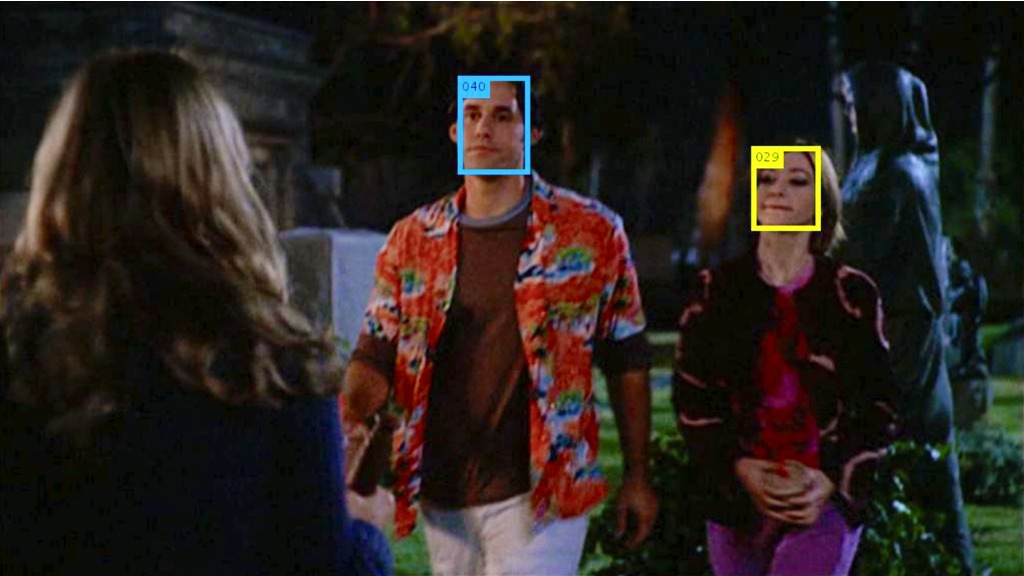} &
      \includegraphics[width=0.2\textwidth]{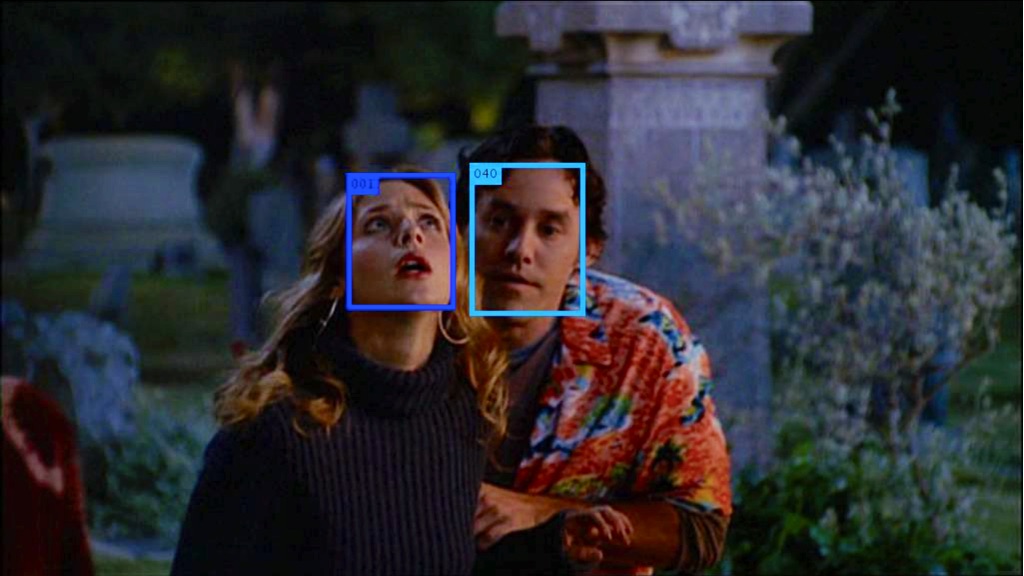} &
      \includegraphics[width=0.2\textwidth]{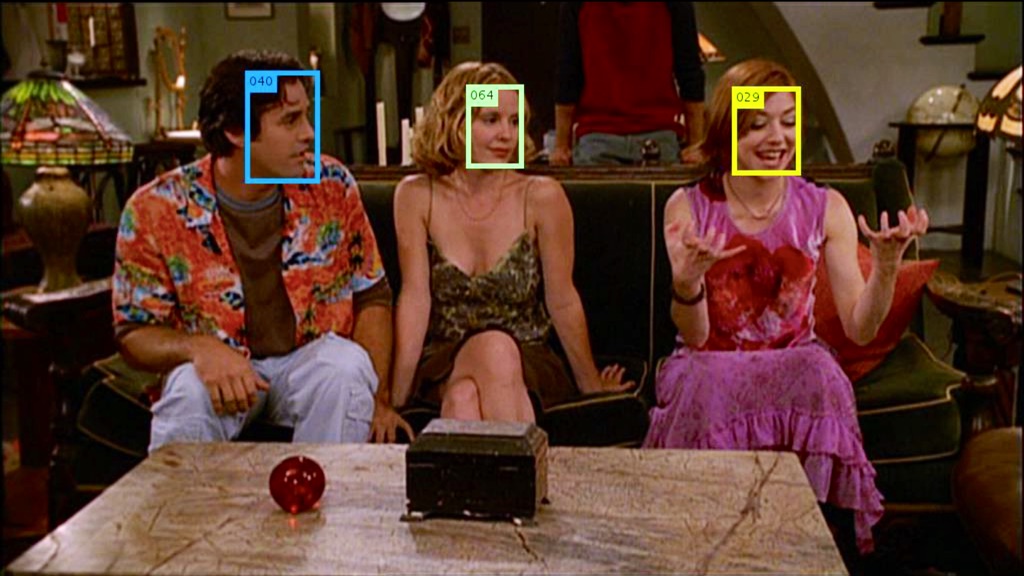}\\
      (f) & (g) & (h) & (i) & (j)\\
      \end{tabular}
    }
    \vspace{-0.3cm}
\caption{{\bf Clustering results from {\em Buffy the Vampire Slayer}}. A failure example can be seen in frame (e), in which the main character Buffy (otherwise in a \textcolor{RoyalPurple}{purple box}) in shown in a \textcolor{RubineRed}{pink box}.}    
\label{fig:cluster_result_buffy}
\end{figure*}

\begin{figure*}[h]
	\centering
	\setlength\tabcolsep{1pt}
    \setlength{\extrarowheight}{-3pt}
    \small{
      \begin{tabular} {c c c c c}
      \includegraphics[width=0.2\textwidth]{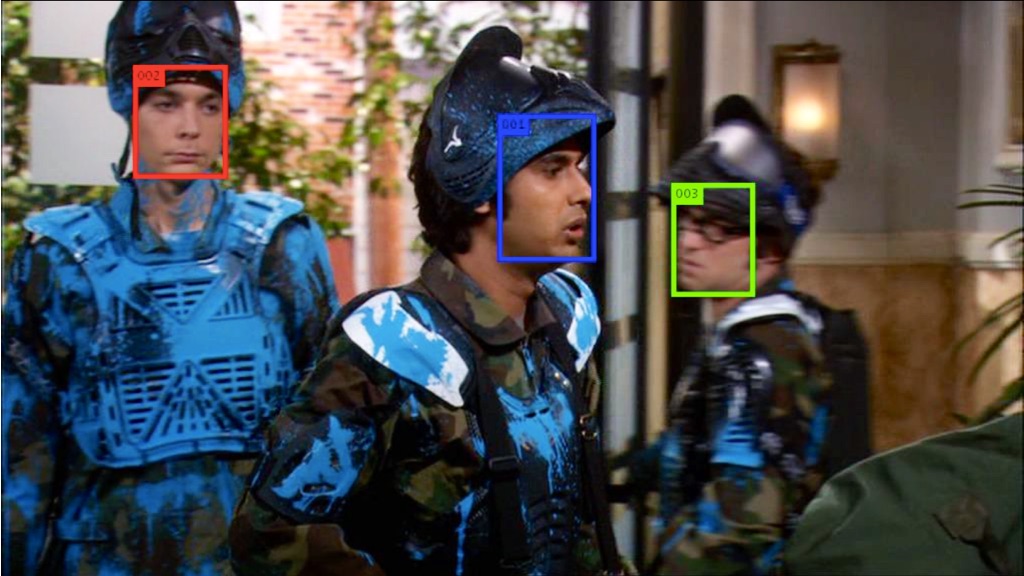} &
      \includegraphics[width=0.2\textwidth]{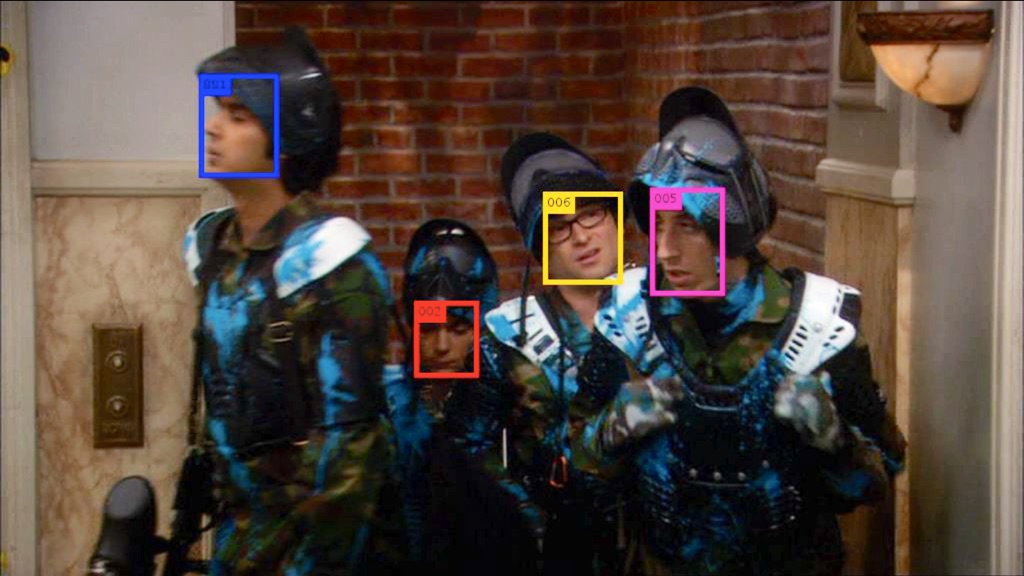} &
      \includegraphics[width=0.2\textwidth]{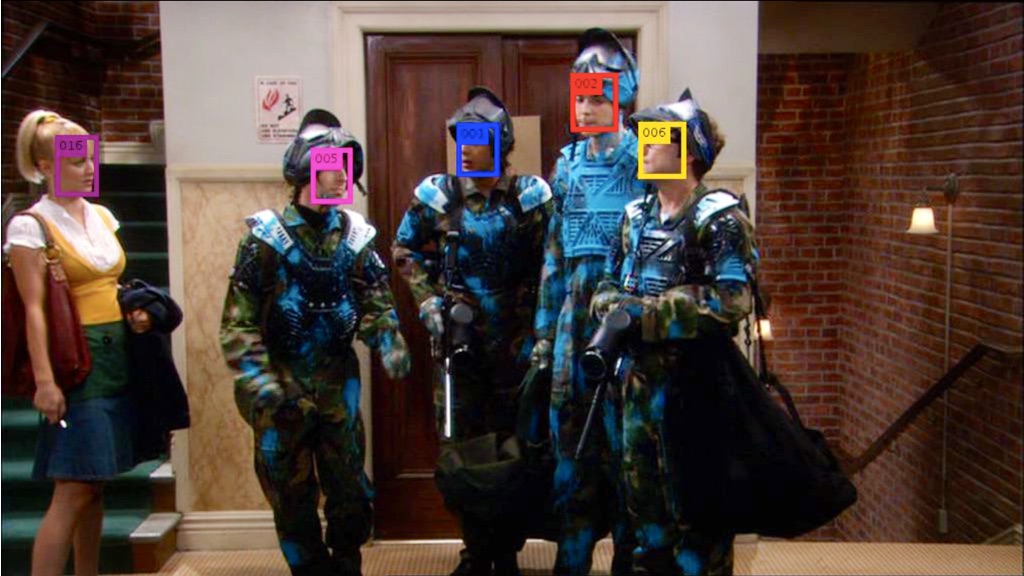} &
      \includegraphics[width=0.2\textwidth]{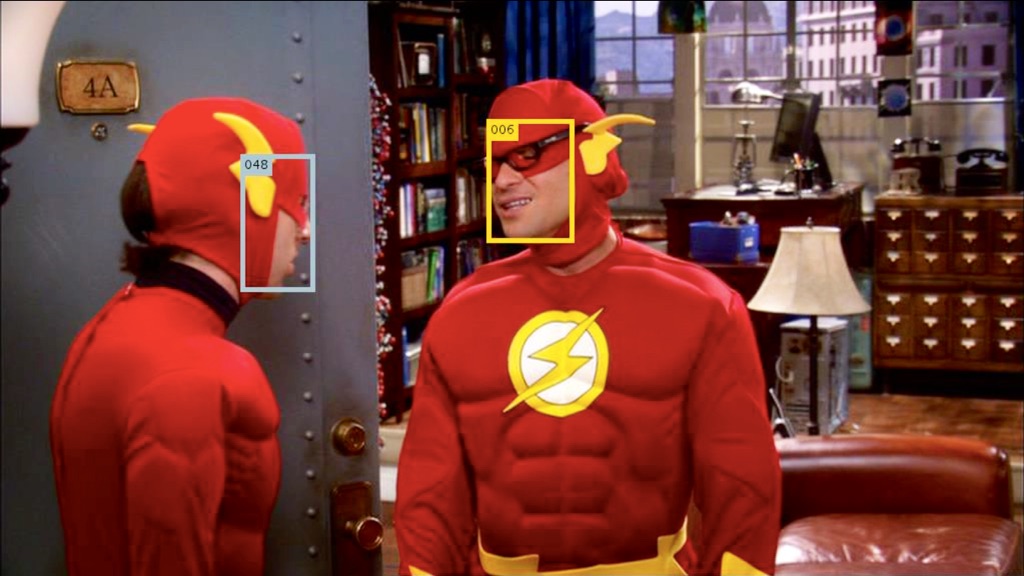} &
      \includegraphics[width=0.2\textwidth]{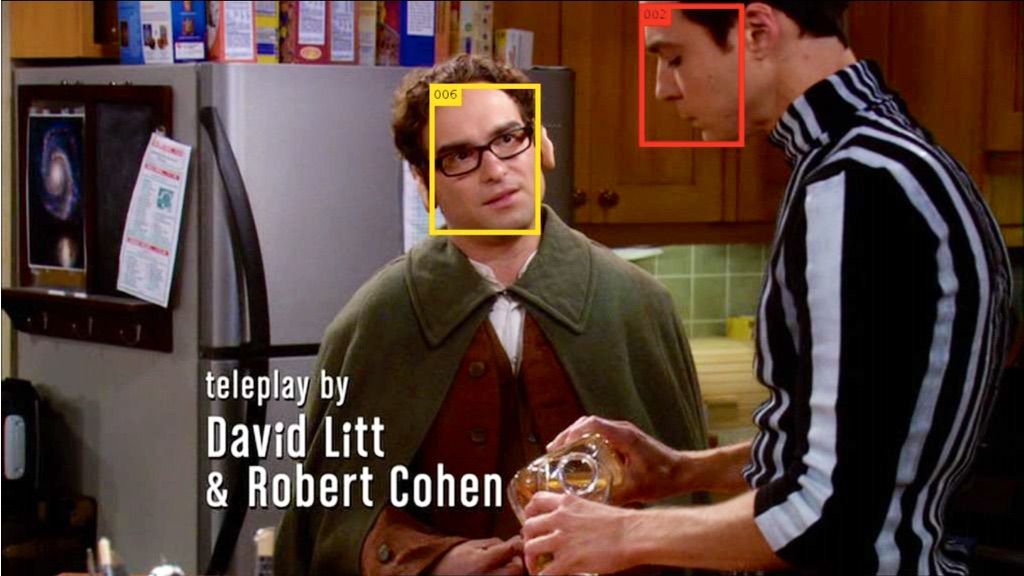} \\
      (a) & (b) & (c) & (d) & (e)\\
      \includegraphics[width=0.2\textwidth]{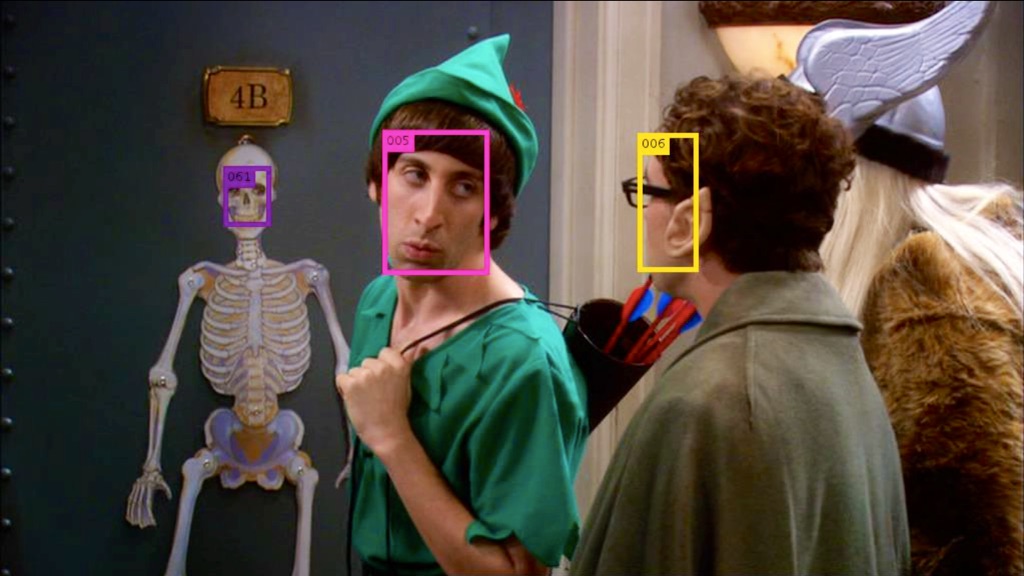} &
      \includegraphics[width=0.2\textwidth]{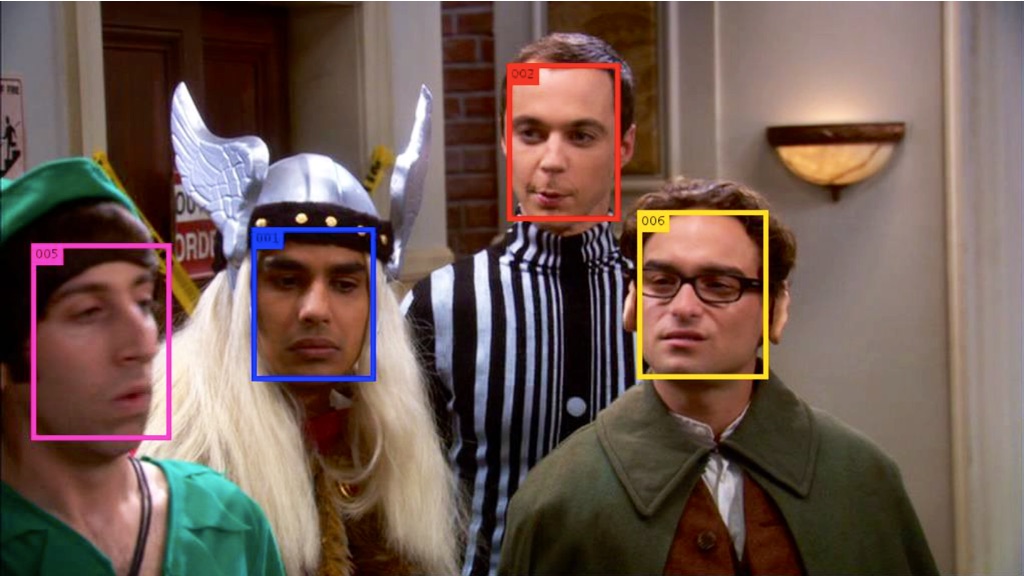} &
      \includegraphics[width=0.2\textwidth]{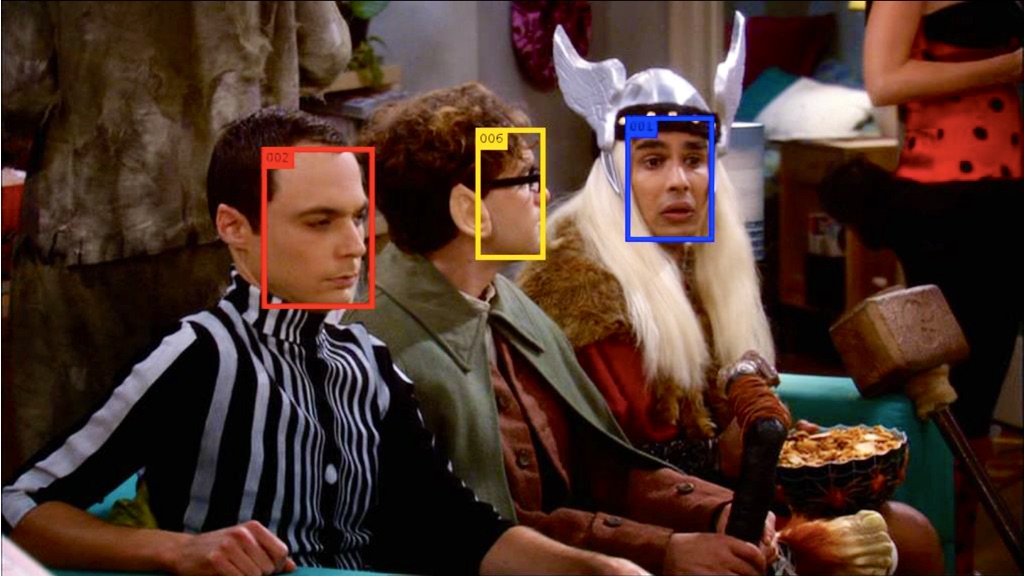} &
      \includegraphics[width=0.2\textwidth]{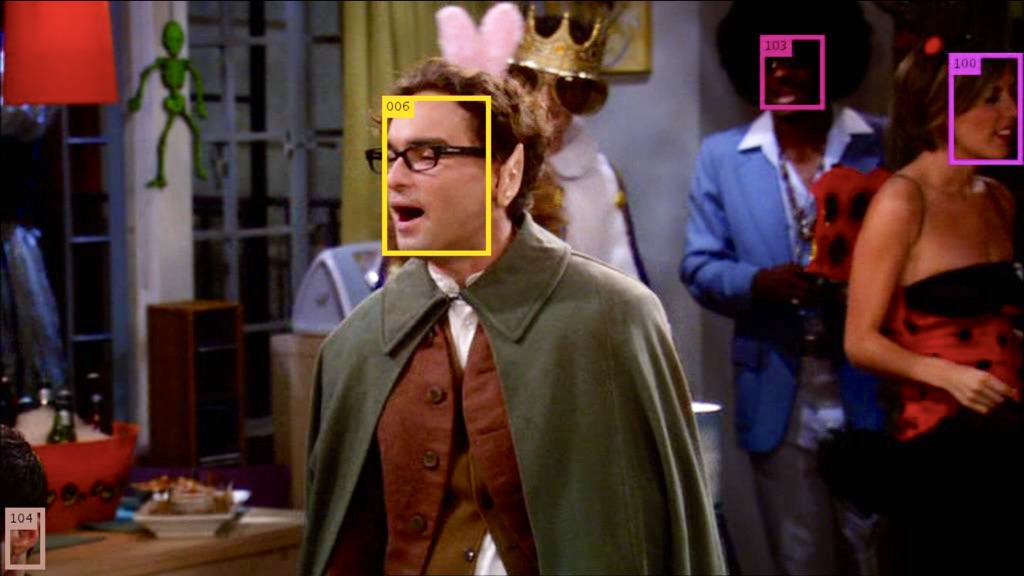} &
      \includegraphics[width=0.2\textwidth]{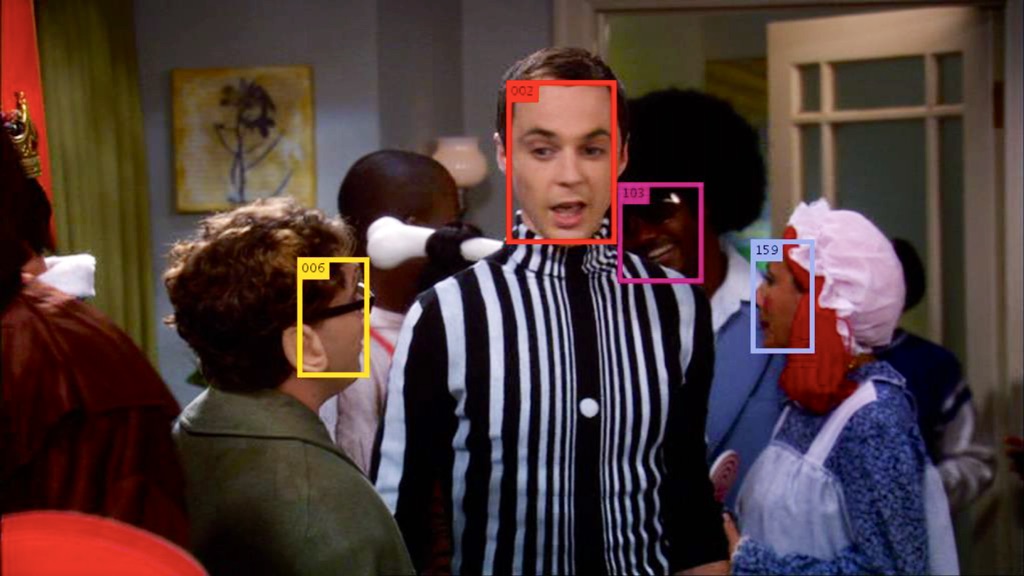} \\
      (f) & (g) & (h) & (i) & (j)\\
      \end{tabular}
    }
    \vspace{-0.3cm}
\caption{{\bf Clustering results from {\em the Big Bang Theory}}. A failure example can be seen in frame (d), in which the main character Howard (otherwise in a \textcolor{Magenta}{magenta box}) in shown in a \textcolor{Gray}{gray box}.}    
\label{fig:cluster_result_bbt}
\end{figure*}

\subsection{Threshold for rank-1 counts}
\label{sec:threshold}

The leftmost column in Table~\ref{tab:clustering_comparison} shows our clustering results when the threshold is set automatically by the validation set. We used LFW as a validation set for BBT, Buffy and Hannah while Hannah was used for LFW. Note that the proposed method is very competitive even when the threshold is automatically set.

\subsection{Comparisons}
\label{sec:comparison}
We can divide other clustering algorithms into two broad categories--link-based clustering algorithms (like ours) that use a different similarity function and clustering algorithms that are not link-based (such as spectral clustering~\cite{shi2000normalized}).
Table~\ref{tab:clustering_comparison} shows the comparisons to various distance functions~\cite{crosswhite2016arxiv,otto2016arxiv,zhu2011cvpr} with our link-based clustering algorithm. L2 shows competitive performance in LFW while the performance drops dramatically when a test set has large pose variations. We also compare against a recent so-called ``template adaptation" method~\cite{crosswhite2016arxiv} which also requires a reference set. It takes 2nd and 3rd place on Buffy and BBT.
In addition, we compare to the rank-order method~\cite{zhu2011cvpr} in two different ways: link-based clustering algorithms using their rank-order distance, and rank-order distance based clustering.

In addition, we compare against several generic clustering algorithms (Affinity Propagation~\cite{frey2007clustering}, DBSCAN~\cite{ester1996density}, Spectral Clustering~\cite{shi2000normalized}, Birch~\cite{zhang1996birch}, KMeans~\cite{sculley2010web}), where L2 distance is used as pairwise metric.
For algorithms that can take as input the similarity matrix (Affinity Propagation, DBSCAN, Spectral Clustering), do-not-link constraints are applied by setting the distance between the corresponding pairs to $\infty$. Note that this is just an approximation, and in general does not guarantee the constraints in the final clustering result (\eg for single-linkage agglomerative clustering, a modified update rule is also needed in Section~\ref{sec:donotlink}). 

Note that all other settings (feature encoding, tracklet generation) are common for all methods. 
In Table~\ref{tab:clustering_comparison}, except for the leftmost column, we report the best $F_{0.5}$ scores using optimal (oracle-supplied) thresholds for (number of clusters, distance). The link-based clustering algorithm with rank-1 counts outperforms the state-of-the-art on all four data sets in $F_{0.5}$ score. Figures~\ref{fig:cluster_result_buffy} and~\ref{fig:cluster_result_bbt} show some clustering results on Buffy and BBT. 

\section{Discussion}
\label{sec:Discussion}
We have presented a system for doing end-to-end clustering in full length videos and movies. In addition to a careful combination of detection and tracking, and a new end-to-end evaluation metric, we have introduced a novel approach to link-based clustering that we call Erd\H{o}s-R\'{e}nyi clustering. 
We demonstrated a method for automatically estimating a good decision threshold for a verification method based on rank-1 counts by estimating the underlying portion of the rank-1 counts distribution due to mismatched pairs. 

This decision threshold was shown to result in good recall at a low false-positive operating point. Such operating points are critical for large clustering problems, since the vast majority of pairs are from different clusters, and false positive links that incorrectly join clusters can have a large negative effect on clustering performance. 

There are several things that could disrupt our algorithm: a) if a high percentage of different pairs are highly similar (e.g. family members), b) if only a small percentage of pairs are “different” (e.g., one cluster contains 90\% of the images), and if “same” pairs lack lots of matching features (e.g., every cluster is a pair of images of the same person under extremely different conditions). 
Nevertheless, we showed excellent results on 3 popular video data sets. Not only do we dominate other methods when thresholds are optimized for clustering, but we outperform other methods even when our thresholds are picked automatically.

\newpage
{\small

\begin{thebibliography}{10}\itemsep=-1pt

\bibitem{barlow1987cerebral}
H.~Barlow.
\newblock Cerebral cortex as model builder.
\newblock In {\em Matters of Intelligence}, pages 395--406. Springer, 1987.

\bibitem{bauml2013cvpr}
M.~Bauml, M.~Tapaswi, and R.~Stiefelhagen.
\newblock Semi-supervised learning with constraints for person identification
  in multimedia data.
\newblock In {\em Proc. {CVPR}}, 2013.

\bibitem{cinbis2011iccv}
R.~G. Cinbis, J.~Verbeek, and C.~Schmid.
\newblock Unsupervised metric learning for face identification in {TV} video.
\newblock In {\em Proc. {ICCV}}, 2011.

\bibitem{crosswhite2016arxiv}
N.~Crosswhite, J.~Byrne, C.~Stauffer, O.~M. Parkhi, Q.~Cao, and A.~Zisserman.
\newblock Template adaptation for face verification and identification.
\newblock In {\em Face and Gesture}, 2017.

\bibitem{davey2008comparison}
S.~J. Davey, M.~G. Rutten, and B.~Cheung.
\newblock A comparison of detection performance for several track-before-detect
  algorithms.
\newblock {\em EURASIP Journal on Advances in Signal Processing}, 2008:41,
  2008.

\bibitem{decann2015biometrics}
B.~DeCann and A.~Ross.
\newblock Modeling errors in a biometric re-identification system.
\newblock {\em IET Biometrics}, 4(4):209--219, 2015.

\bibitem{ErdosRenyiGraphs}
P.~Erd\H{o}s and A.~R\'{e}nyi.
\newblock On the evolution of random graphs.
\newblock {\em Publications of the Mathematical Institute of the Hungarian
  Academy of Sciences}, 5:17--61, 1960.

\bibitem{ester1996density}
M.~Ester, H.-P. Kriegel, J.~Sander, and X.~Xu.
\newblock A density-based algorithm for discovering clusters in large spatial
  databases with noise.
\newblock {\em KDD}, 96(34):226--231, 1996.

\bibitem{everingham2006cvpr}
M.~Everingham, J.~Sivic, and A.~Zisserman.
\newblock "{Hello! My name is... Buffy}" {Automatic} naming of characters in
  {TV} video.
\newblock In {\em Proc. {BMVC}}, 2006.

\bibitem{forney1973viterbi}
G.~D. Forney.
\newblock The {Viterbi} algorithm.
\newblock {\em Proceedings of the IEEE}, 61(3):268--278, 1973.

\bibitem{frey2007clustering}
B.~J. Frey and D.~Dueck.
\newblock Clustering by passing messages between data points.
\newblock {\em Science}, 315(5814):972--976, 2007.

\bibitem{gyaourova2012tifs}
A.~Gyaourova and A.~Ross.
\newblock Index codes for multibiometric pattern retrieval.
\newblock {\em IEEE Transactions on Information Forensics and Security (TIFS)},
  7(2):518--529, April 2012.

\bibitem{haurilet2016cvpr}
M.-L. Haurilet, M.~Tapaswi, Z.~Al-Halah, and R.~Stiefelhagen.
\newblock Naming {TV} characters by watching and analyzing dialogs.
\newblock In {\em Proc. {CVPR}}, 2016.

\bibitem{LFW}
G.~B. Huang, M.~Mattar, T.~Berg, and E.~Learned-Miller.
\newblock Labeled faces in the wild: A database for studying face recognition
  in unconstrained environments.
\newblock In {\em The Workshop on Faces in Real-Life Images at ECCV}, 2008.

\bibitem{jiang2017}
H.~Jiang and E.~Learned-Miller.
\newblock Face detection with the {Faster R-CNN}.
\newblock In {\em Face and Gesture}, 2017.

\bibitem{kuhn1955hungarian}
H.~W. Kuhn.
\newblock The hungarian method for the assignment problem.
\newblock {\em Naval research logistics quarterly}, 2(1-2):83--97, 1955.

\bibitem{li2008icml}
Z.~Li, J.~Liu, and X.~Tang.
\newblock Pairwise constraint propagation by semidefinite programming for
  semi-supervised classification.
\newblock In {\em Proc. {ICML}}, 2008.

\bibitem{lisanti2015pami}
G.~Lisanti, I.~Masi, A.~D. Bagdanov, and A.~D. Bimbo.
\newblock Person re-identification by iterative re-weighted sparse ranking.
\newblock {\em TPAMI}, 37(8):1629--1642, August 2015.

\bibitem{lu2007neural}
Z.~Lu and T.~K. Leen.
\newblock Penalized probabilistic clustering.
\newblock {\em Neural Computation}, 19(6):1528--1567, 2007.

\bibitem{MOT16}
A.~Milan, L.~Leal-Taix\'{e}, I.~Reid, S.~Roth, and K.~Schindler.
\newblock {MOT}16: {A} benchmark for multi-object tracking.
\newblock {\em arXiv:1603.00831 [cs]}, Mar. 2016.
\newblock arXiv: 1603.00831.

\bibitem{miyamoto2010fuzz}
S.~Miyamoto and A.~Terami.
\newblock Semi-supervised agglomerative hierarchical clustering algorithms with
  pairwise constraints.
\newblock In {\em Fuzzy Systems (FUZZ)}, pages 1--6. IEEE, 2010.

\bibitem{murtagh2012algorithms}
F.~Murtagh and P.~Contreras.
\newblock Algorithms for hierarchical clustering: an overview.
\newblock {\em Wiley Interdisciplinary Reviews: Data Mining and Knowledge
  Discovery}, 2(1):86--97, 2012.

\bibitem{otto2016arxiv}
C.~Otto, D.~Wang, and A.~K. Jain.
\newblock Clustering millions of faces by identity.
\newblock {\em TPAMI}, Mar. 2017.

\bibitem{Hannah}
A.~Ozerov, J.-R. Vigouroux, L.~Chevallier, and P.~P{\'e}rez.
\newblock On evaluating face tracks in movies.
\newblock In {\em Proc. {ICIP}}, 2013.

\bibitem{ozerov2013icip}
A.~Ozerov, J.-R. Vigouroux, L.~Chevallier, and P.~P{\'e}rez.
\newblock On evaluating face tracks in movies.
\newblock In {\em Proc. {ICIP}}. IEEE, 2013.

\bibitem{parkhi2015bmvc}
O.~M. Parkhi, A.~Vedaldi, and A.~Zisserman.
\newblock Deep face recognition.
\newblock In {\em bmvc}, 2015.

\bibitem{TV}
A.~Patron-Perez, M.~Marszałek, A.~Zisserman, and I.~D. Reid.
\newblock High five: Recognising human interactions in tv shows.
\newblock In {\em Proc. {BMVC}}, 2010.

\bibitem{FasterRCNN}
S.~Ren, K.~He, R.~B. Girshick, and J.~Sun.
\newblock Faster {R-CNN}: towards real-time object detection with region
  proposal networks.
\newblock In {\em Proc. {NIPS}}, 2015.

\bibitem{roth2012icpr}
M.~Roth, M.~Bauml, R.~Nevatia, and R.~Stiefelhagen.
\newblock Robust multi-pose face tracking by multi-stage tracklet association.
\newblock In {\em Proc. {ICPR}}, 2012.

\bibitem{sculley2010web}
D.~Sculley.
\newblock Web-scale k-means clustering.
\newblock In {\em Proc. {WWW}}, pages 1177--1178. ACM, 2010.

\bibitem{lara2012cvpr}
L.~Sevilla-Lara and E.~Learned-Miller.
\newblock Distribution fields for tracking.
\newblock In {\em Proc. {CVPR}}, 2012.

\bibitem{shental2004nips}
N.~Shental, A.~Bar-Hillel, T.~Hertz, and D.~Weinshall.
\newblock Computing {Gaussian} mixture models with {EM} using equivalence
  constraints.
\newblock In {\em Proc. {NIPS}}, 2004.

\bibitem{shi2000normalized}
J.~Shi and J.~Malik.
\newblock Normalized cuts and image segmentation.
\newblock {\em TPAMI}, 22(8):888--905, 2000.

\bibitem{sibson1973slink}
R.~Sibson.
\newblock {SLINK}: an optimally efficient algorithm for the single-link cluster
  method.
\newblock {\em The computer journal}, 16(1):30--34, 1973.

\bibitem{tapaswi2012cvpr}
M.~Tapaswi, M.~Bauml, and R.~Stiefelhagen.
\newblock {"Knock! Knock! Who is it?"} {Probabilistic} person identification in
  {TV} series.
\newblock In {\em Proc. {CVPR}}, 2012.

\bibitem{tapaswi2014icip}
M.~Tapaswi, C.~C. Corez, M.~Bauml, H.~K. Ekenel, and R.~Stiefelhagen.
\newblock Cleaning up after a face tracker: False positive removal.
\newblock In {\em Proc. {ICIP}}, 2014.

\bibitem{tapaswi2015icvgip}
M.~Tapaswi, O.~M. Parkhi, E.~Rahtu, E.~Sommerlade, R.~Stiefelhagen, and
  A.~Zisserman.
\newblock Total cluster: A person agnostic clustering method for broadcast
  videos.
\newblock In {\em ICVGIP}, 2014.

\bibitem{wagstaff2001icml}
K.~Wagstaff, C.~Cardie, S.~Rogers, S.~Schr{\"o}dl, et~al.
\newblock Constrained k-means clustering with background knowledge.
\newblock In {\em Proc. {ICML}}, 2001.

\bibitem{wu2013iccv}
B.~Wu, S.~Lyu, B.-G. Hu, and Q.~Ji.
\newblock Simultaneous clustering and tracklet linking for multi-face tracking
  in videos.
\newblock In {\em Proc. {ICCV}}, 2013.

\bibitem{wu2013cvpr}
B.~Wu, Y.~Zhang, B.-G. Hu, and Q.~Ji.
\newblock Constrained clustering and its application to face clustering in
  videos.
\newblock In {\em Proc. {CVPR}}, 2013.

\bibitem{yang2015wider}
S.~Yang, P.~Luo, C.~C. Loy, and X.~Tang.
\newblock Wider face: A face detection benchmark.
\newblock In {\em CVPR}, 2016.

\bibitem{zhang1996birch}
T.~Zhang, R.~Ramakrishnan, and M.~Livny.
\newblock Birch: an efficient data clustering method for very large databases.
\newblock In {\em SIGMOD}. ACM, 1996.

\bibitem{zhang2016eccv}
Z.~Zhang, P.~Luo, C.~C. Loy, and X.~Tang.
\newblock Joint face representation adaptation and clustering in videos.
\newblock In {\em Proc. {ECCV}}, 2016.

\bibitem{zheng2016arxiv}
L.~Zheng, Y.~Yang, and A.~G. Hauptman.
\newblock Person re-identification: Past, present and future.
\newblock {\em arXiv}, Oct. 2016.

\bibitem{zhu2011cvpr}
C.~Zhu, F.~Wen, and J.~Sun.
\newblock A rank-order distance based clustering algorithm for face tagging.
\newblock In {\em Proc. {CVPR}}, 2011.

\end{thebibliography}

}

\begin{appendices}

\begin{table*}[!tb]
\centering
	\vspace{-1.2cm}
	\captionof{table}{Clustering performance comparisons on various data sets. The leftmost shows our \textbf{rank1count} by setting a threshold automatically. For the rest of the columns, we show f-scores using optimal (oracle-supplied) thresholds. (\textcolor{red}{\textbf{1st place}},\textcolor{orange}{\textbf{2nd place}},\textcolor{ForestGreen}{\textbf{3rd place}}).}
    \label{tab:clustering_comparison}

    \scalebox{0.8}{
	\begin{tabular}{ c | c | c | p{1.2cm}| p{1.2cm} p{1.2cm} p{1.2cm} p{1.2cm} | p{1.2cm} p{1.2cm} p{1.2cm} p{1.2cm} p{1.2cm} p{1.2cm} } 
        \Xhline{1.5pt}
        \multicolumn{3}{ c |}{ }
        		& \multicolumn{5}{ c |}{ Verification system + Link-based clustering algorithm } 
                & \multicolumn{6}{ c  }{ Other clustering algorithms }
                \\
        \cmidrule{4-14}
		\multicolumn{3}{ c |}{Test set}
        		& \textbf{ \rotatebox[origin=l]{90}{\pbox{5cm}{Rank-1 Count\\(automatic threshold)}} }
        		& \textbf{ \rotatebox[origin=l]{90}{\pbox{5cm}{Rank-1 Count}} }
                & \rotatebox[origin=l]{90}{\pbox{5cm}{L2}}
                & \rotatebox[origin=l]{90}{\pbox{5cm}{Template\\Adaptation~\cite{crosswhite2016arxiv}}} 
                & \rotatebox[origin=l]{90}{\pbox{5cm}{Rank-Order\\Distance~\cite{zhu2011cvpr}}}
                & \rotatebox[origin=l]{90}{\pbox{5cm}{Rank-Order Distance\\based Clustering~\cite{zhu2011cvpr}}}
                & \rotatebox[origin=l]{90}{\pbox{5cm}{Affinity\\Propagation~\cite{frey2007clustering}}} 
                & \rotatebox[origin=l]{90}{\pbox{5cm}{DBSCAN~\cite{ester1996density}}} 
                & \rotatebox[origin=l]{90}{\pbox{5cm}{Spectral\\Clustering~\cite{shi2000normalized}}} 
                & \rotatebox[origin=l]{90}{\pbox{5cm}{Birch~\cite{zhang1996birch}}} 
                & \rotatebox[origin=l]{90}{\pbox{5cm}{MiniBatch\\KMeans~\cite{sculley2010web}}} 
                \\
\Xhline{1.5pt}
\multirow{12}{*}{Video} 
& \multirow{6}{*}{\pbox{1cm}{BBT s01 \cite{bauml2013cvpr}}}  
	& e01 & .7145 & .7225 & \textcolor{orange}{\textbf{.7386}} & .7170 & \textcolor{red}{\textbf{.8064}} & \textcolor{ForestGreen}{\textbf{.7278}} & .1707 & .4137 & .6884 & .3776 & .2166 \\ 
    & & e02 & .7414 & \textcolor{red}{\textbf{.7671}} & \textcolor{orange}{\textbf{.7561}} & \textcolor{ForestGreen}{\textbf{.7520}} & .7154 & .6537 & .1593 & .3216 & .6147 & .2337 & .2018 \\ 
    & & e03 & \textcolor{orange}{\textbf{.8428}} & \textcolor{red}{\textbf{.8552}} & \textcolor{ForestGreen}{\textbf{.8329}} & .8192 & .6660 & .6367 & .2130 & .2985 & .6578 & .2366 & .2131 \\ 
    & & e04 & \textcolor{ForestGreen}{\textbf{.7602}} & \textcolor{red}{\textbf{.7690}} & .7151 & \textcolor{orange}{\textbf{.7687}} & .6364 & .7001 & .2118 & .2886 & .6520 & .2156 & .1847 \\ 
    & & e05 & \textcolor{orange}{\textbf{.8217}} & \textcolor{red}{\textbf{.8250}} & .7420 & \textcolor{ForestGreen}{\textbf{.7858}} & .6330 & .7035 & .2335 & .2444 & .5980 & .1812 & .2120 \\ 
    & & e06 & \textcolor{orange}{\textbf{.7563}} & \textcolor{red}{\textbf{.7578}} & .6342 & \textcolor{ForestGreen}{\textbf{.7247}} & .5577 & .5588 & .1615 & .1948 & .5806 & .1511 & .1387 \\ 
\cmidrule{2-14}
& \multirow{6}{*}{\pbox{1cm}{Buffy s05 \cite{bauml2013cvpr}}}
	& e01 & \textcolor{ForestGreen}{\textbf{.6634}} & \textcolor{red}{\textbf{.6938}} & .4950 & \textcolor{orange}{\textbf{.6902}} & .3819 & .5935 & .1711 & .1755 & .5762 & .1439 & .1285 \\ 
  	& & e02 & .5582 & \textcolor{red}{\textbf{.6645}} & .3315 & .5452 & .2800 & \textcolor{ForestGreen}{\textbf{.5837}} & .1705 & .1185 & \textcolor{orange}{\textbf{.5892}} & .1151 & .1087 \\ 
  	& & e03 & \textcolor{ForestGreen}{\textbf{.5378}} & \textcolor{orange}{\textbf{.5479}} & .3735 & \textcolor{red}{\textbf{.5569}} & .2390 & .4595 & .1346 & .1322 & .4566 & .1077 & .1063 \\ 
  	& & e04 & .4203 & \textcolor{ForestGreen}{\textbf{.4859}} & .3523 & .4549 & .3049 & \textcolor{orange}{\textbf{.5171}} & .1643 & .1445 & \textcolor{red}{\textbf{.5273}} & .1187 & .1179 \\ 
  	& & e05 & \textcolor{ForestGreen}{\textbf{.6235}} & \textcolor{red}{\textbf{.6952}} & .5064 & \textcolor{orange}{\textbf{.6739}} & .3073 & .5640 & .1435 & .1740 & .5540 & .1390 & .1251 \\ 
  	& & e06 & \textcolor{orange}{\textbf{.5932}} & \textcolor{red}{\textbf{.6923}} & .3001 & \textcolor{ForestGreen}{\textbf{.5856}} & .2807 & .5455 & .1765 & .1009 & .5071 & .1041 & .0995 \\ 
\cmidrule{2-14}
& \multicolumn{2}{ c |}{Hannah~\cite{Hannah}} 
& \textcolor{orange}{\textbf{.6436}} & \textcolor{red}{\textbf{.6813}} & .2581 & .3620 & \textcolor{ForestGreen}{\textbf{.4123}} & .3955 & .1886 & .1230 & .3344 & .1240 & .1052 \\ 
\midrule                                
\multirow{1}{*}{Image} & \multicolumn{2}{ c |}{LFW~\cite{LFW}} 
& \textcolor{orange}{\textbf{.8532}} & \textcolor{red}{\textbf{.8943}} & \textcolor{ForestGreen}{\textbf{.8498}} & .3735 & .5989 & .5812 & .3197 & .0117 & .2538 & .4520 & .3133 \\ 
\Xhline{1.5pt}      
\end{tabular}
}

\end{table*}

\begin{table*}[!tb]
\centering
\caption{Clustering performance comparisons evaluated on traditional measures. (\textcolor{red}{\textbf{1st place}},\textcolor{orange}{\textbf{2nd place}},\textcolor{ForestGreen}{\textbf{3rd place}}).}
\label{tab:clustering_comparison2}
\scalebox{0.8}{
	\begin{tabular}{ c | c | c | p{1.2cm}| p{1.2cm} p{1.2cm} p{1.2cm} p{1.2cm} | p{1.2cm} p{1.2cm} p{1.2cm} p{1.2cm} p{1.2cm} p{1.2cm} } 
        \Xhline{1.5pt}
        \multicolumn{3}{ c |}{ }
        		& \multicolumn{5}{ c |}{ Verification system + Link-based clustering algorithm } 
                & \multicolumn{6}{ c  }{ Other clustering algorithms }
                \\
        \cmidrule{4-14}
		\multicolumn{3}{ c |}{Test set}
        		& \textbf{ \rotatebox[origin=l]{90}{\pbox{5cm}{Rank-1 Count\\(automatic threshold)}} }
        		& \textbf{ \rotatebox[origin=l]{90}{\pbox{5cm}{Rank-1 Count}} }
                & \rotatebox[origin=l]{90}{\pbox{5cm}{L2}}
                & \rotatebox[origin=l]{90}{\pbox{5cm}{Template\\Adaptation~\cite{crosswhite2016arxiv}}} 
                & \rotatebox[origin=l]{90}{\pbox{5cm}{Rank-Order\\Distance~\cite{zhu2011cvpr}}}
                & \rotatebox[origin=l]{90}{\pbox{5cm}{Rank-Order Distance\\based Clustering~\cite{zhu2011cvpr}}}
                & \rotatebox[origin=l]{90}{\pbox{5cm}{Affinity\\Propagation~\cite{frey2007clustering}}} 
                & \rotatebox[origin=l]{90}{\pbox{5cm}{DBSCAN~\cite{ester1996density}}} 
                & \rotatebox[origin=l]{90}{\pbox{5cm}{Spectral\\Clustering~\cite{shi2000normalized}}} 
                & \rotatebox[origin=l]{90}{\pbox{5cm}{Birch~\cite{zhang1996birch}}} 
                & \rotatebox[origin=l]{90}{\pbox{5cm}{MiniBatch\\KMeans~\cite{sculley2010web}}} 
                \\
\Xhline{1.5pt}
\multirow{12}{*}{Video}
& \multirow{6}{*}{\pbox{1cm}{BBT s01 \cite{bauml2013cvpr}}}  
	& e01 & .8226 & .8486 & \textcolor{ForestGreen}{\textbf{.8613}} & .8364 & \textcolor{red}{\textbf{.9669}} & \textcolor{orange}{\textbf{.8732}} & .2108 & .4880 & .8193 & .4572 & .2666 \\ 
    & & e02 & .9289 & \textcolor{red}{\textbf{.9726}} & \textcolor{orange}{\textbf{.9550}} & \textcolor{ForestGreen}{\textbf{.9502}} & .9046 & .8456 & .2034 & .4094 & .7785 & .2997 & .2584 \\ 
    & & e03 & \textcolor{ForestGreen}{\textbf{.9664}} & \textcolor{red}{\textbf{.9908}} & \textcolor{orange}{\textbf{.9903}} & .9299 & .7873 & .7751 & .2437 & .4404 & .7810 & .3158 & .2775 \\ 
    & & e04 & \textcolor{ForestGreen}{\textbf{.8985}} & \textcolor{red}{\textbf{.9188}} & .8638 & \textcolor{orange}{\textbf{.9107}} & .8040 & .8387 & .2592 & .3831 & .8207 & .2863 & .2385 \\ 
    & & e05 & \textcolor{orange}{\textbf{.9769}} & \textcolor{red}{\textbf{.9940}} & .9215 & \textcolor{ForestGreen}{\textbf{.9275}} & .7879 & .8067 & .2600 & .3866 & .7638 & .2713 & .3354 \\ 
    & & e06 & \textcolor{orange}{\textbf{.9795}} & \textcolor{red}{\textbf{.9876}} & .8605 & \textcolor{ForestGreen}{\textbf{.9644}} & .7828 & .7041 & .2085 & .3930 & .8408 & .2859 & .2538 \\ 
\cmidrule{2-14}

& \multirow{6}{*}{\pbox{1cm}{Buffy s05 \cite{bauml2013cvpr}}}
	& e01 & \textcolor{ForestGreen}{\textbf{.8487}} & \textcolor{red}{\textbf{.8727}} & .7016 & \textcolor{orange}{\textbf{.8626}} & .5447 & .7249 & .1997 & .2968 & .7512 & .2123 & .1835 \\ 
  	& & e02 & \textcolor{ForestGreen}{\textbf{.6737}} & \textcolor{orange}{\textbf{.7665}} & .4904 & .6730 & .4104 & .6384 & .1883 & .2386 & \textcolor{red}{\textbf{.8196}} & .1932 & .1764 \\ 
  	& & e03 & \textcolor{ForestGreen}{\textbf{.6872}} & \textcolor{orange}{\textbf{.7159}} & .5492 & \textcolor{red}{\textbf{.7404}} & .3731 & .6094 & .1538 & .2494 & .6392 & .1725 & .1638 \\ 
  	& & e04 & .5496 & \textcolor{ForestGreen}{\textbf{.5847}} & .4983 & .5586 & .4570 & \textcolor{orange}{\textbf{.6789}} & .1854 & .2603 & \textcolor{red}{\textbf{.6943}} & .1854 & .1733 \\ 
  	& & e05 & \textcolor{orange}{\textbf{.8205}} & \textcolor{red}{\textbf{.8301}} & .6682 & \textcolor{ForestGreen}{\textbf{.8173}} & .4751 & .6663 & .1573 & .3536 & .7686 & .2279 & .2019 \\ 
  	& & e06 & \textcolor{orange}{\textbf{.7509}} & \textcolor{red}{\textbf{.8555}} & .4281 & \textcolor{ForestGreen}{\textbf{.7083}} & .4489 & .6224 & .2071 & .1925 & .7188 & .1727 & .1583 \\ 
\cmidrule{2-14}

& \multicolumn{2}{ c |}{Hannah~\cite{Hannah}} 
& \textcolor{orange}{\textbf{.7634}} & \textcolor{red}{\textbf{.8081}} & .3526 & .4766 & \textcolor{ForestGreen}{\textbf{.5249}} & .4955 & .2272 & .1723 & .4227 & .1648 & .1387 \\ 
\midrule           

\multirow{1}{*}{Image} & \multicolumn{2}{ c |}{LFW~\cite{LFW}} 
& \textcolor{orange}{\textbf{.8532}} & \textcolor{red}{\textbf{.8943}} & \textcolor{ForestGreen}{\textbf{.8498}} & .3735 & .5989 & .5812 & .3197 & .0117 & .2538 & .4520 & .3133 \\ 
\Xhline{1.5pt}      
\end{tabular}
}
\end{table*}

\section{Performance Comparisons}
\label{sec:performance_comparisons}
In Table~\ref{tab:clustering_comparison}, except for the leftmost column of results, we report the best $F_{0.5}$ scores using optimal (oracle-supplied) thresholds for both the distance threshold (a parameter that is part of all of the algorithms) and the number of clusters (a parameter required by a subset of the algorithms, such as k-nearest neighbors). \textbf{The comparison shows that the proposed link-based clustering algorithm with rank-1 counts outperforms the state-of-the-art on all four data sets in $F_{0.5}$ score}. Unlike other clustering algorithms, our proposed approach can scale from small clustering problems (5-8 subjects in BBT) to large clustering problems (5730 subjects in LFW).

In Table~\ref{tab:clustering_comparison2}, we also report traditional measures (pairwise precision, pairwise recall, and F-measure) on the subset of true positive tracklets that are given to each algorithm. 

\end{appendices}

\end{document}